\documentclass[journal]{IEEEtran}

\usepackage{amssymb}
\usepackage{hyperref}
\usepackage{siunitx} 
\DeclareSIUnit\px{px}

\usepackage{newunicodechar}
\usepackage{pifont}
\newunicodechar{✔}{\ding{52}}
\newunicodechar{✗}{\ding{55}}

\usepackage{float}

\usepackage[labelformat=simple]{subcaption}
\usepackage{graphicx}

\usepackage{hhline}

\usepackage{amsmath}
\usepackage{array}

\usepackage{tikz}
\newcommand{\votmedal}[3]{\tikz[baseline=(char.base)]{\node[rounded corners=2pt,fill=#1,draw=#2,inner sep=1.5pt] (char) {#3};}}

\definecolor{gold}{HTML}{FBF2D2}
\definecolor{silver}{HTML}{DDDDDD}
\definecolor{bronze}{HTML}{EED2B8}

\definecolor{goldD}{HTML}{D9AE13}
\definecolor{silverD}{HTML}{909090}
\definecolor{bronzeD}{HTML}{9A5F26}

\newcommand{\medal}[2]{
    \ifcase#1\or
      {\votmedal{gold}{goldD}{\textbf{#2}}}
    \or 
      {\votmedal{silver}{silverD}{#2}}
    \or 
      {\votmedal{bronze}{bronzeD}{#2}}
    \else 
       #2
    \fi\ignorespaces
}

\begin{document}

\title{MULTIAQUA: A multimodal maritime dataset and robust training strategies for multimodal semantic segmentation}

%
%
%

\author{Jon Muhovi\v{c}, Janez Per\v{s}}

\maketitle

\begin{abstract}

Unmanned surface vehicles can encounter a number of varied visual circumstances during operation, some of which can be very difficult to interpret. While most cases can be solved only using color camera images, some weather and lighting conditions require additional information. To expand the available maritime data, we present a novel multimodal maritime dataset MULTIAQUA (Multimodal Aquatic Dataset). Our dataset contains synchronized, calibrated and annotated data captured by sensors of different modalities, such as RGB, thermal, IR, LIDAR, etc. The dataset is aimed at developing supervised methods that can extract useful information from these modalities in order to provide a high quality of scene interpretation regardless of potentially poor visibility conditions. To illustrate the benefits of the proposed dataset, we evaluate several multimodal methods on our difficult nighttime test set. We present training approaches that enable multimodal methods to be trained in a more robust way, thus enabling them to retain reliable performance even in near-complete darkness. Our approach allows for training a robust deep neural network only using daytime images, thus significantly simplifying data acquisition, annotation, and the training process. The dataset is available at \url{https://lmi.fe.uni-lj.si/en/MULTIAQUA}.
\end{abstract}



%
\IEEEpeerreviewmaketitle

\section{Introduction}
%
%
%
%

The difficulty of maritime scene interpretation can be very varied, depending on the visual conditions. As opposed to navigating roads in a ground vehicle where the drivable areas and obstacles are limited, the circumstances can be very varied when observing water surfaces. The appearance changes due to waves and time of day can be significant, especially when direct sunlight is involved. This causes glitter on the water surface as well as a large increase in image contrast, which can effectively hide important parts of the scene inside especially bright or especially dark areas of the image. Similarly, foggy or poorly lit scenes can reduce the contrast to a point where determining important or dangerous parts of the scene again becomes difficult. To alleviate such issues, additional sensors with different modalities can be used as both a source of redundancy, and also as a way to extract information from scenes where purely visible spectrum sensors perform poorly.

\begin{figure*}[ht]
    \centering
    \def\twidth{0.44}

    \bgroup
    \def\arraystretch{1.5}
    \begin{tabular}{cc}
    
    \frame{\includegraphics[width=\twidth\linewidth]{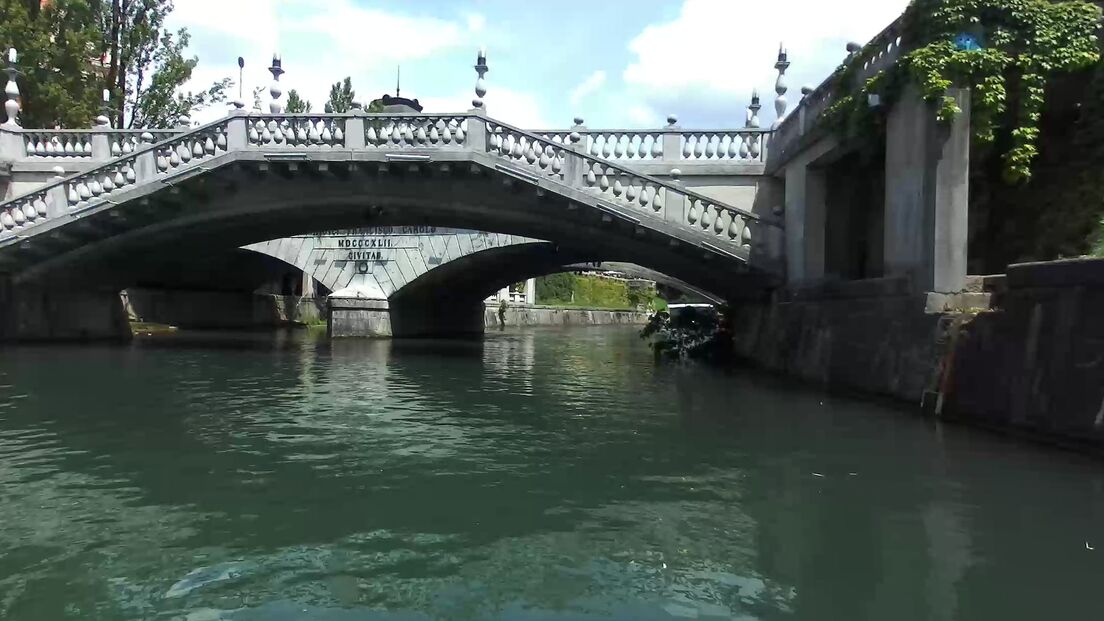}} &
    \frame{\includegraphics[width=\twidth\linewidth]{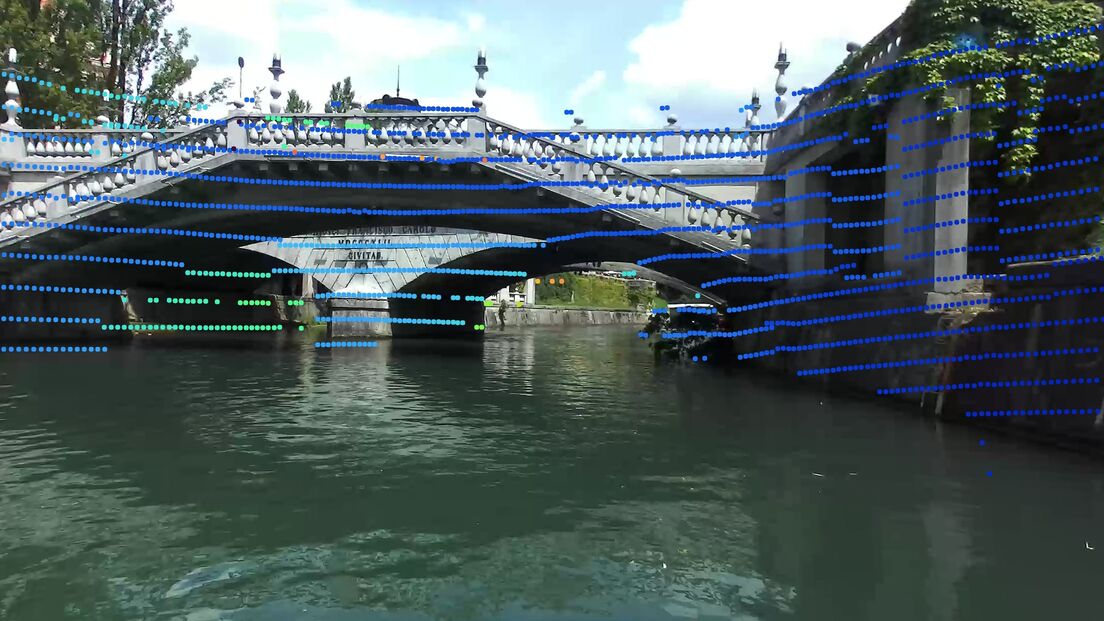}}\\

    \vspace{0.25cm}

    \frame{\includegraphics[width=\twidth\linewidth]{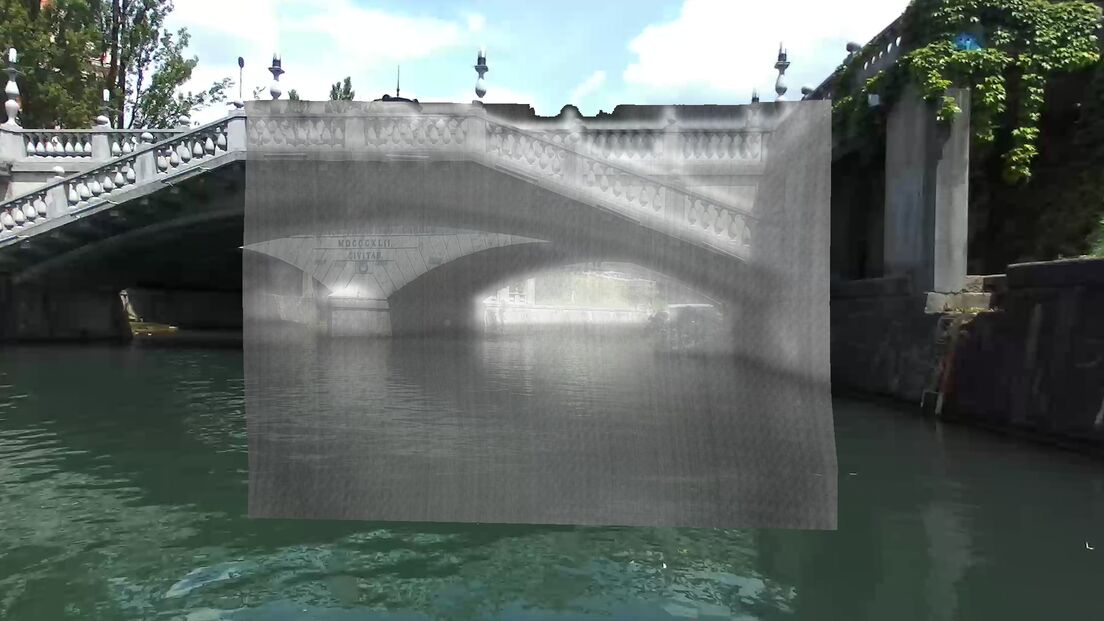}} &
    \frame{\includegraphics[width=\twidth\linewidth]{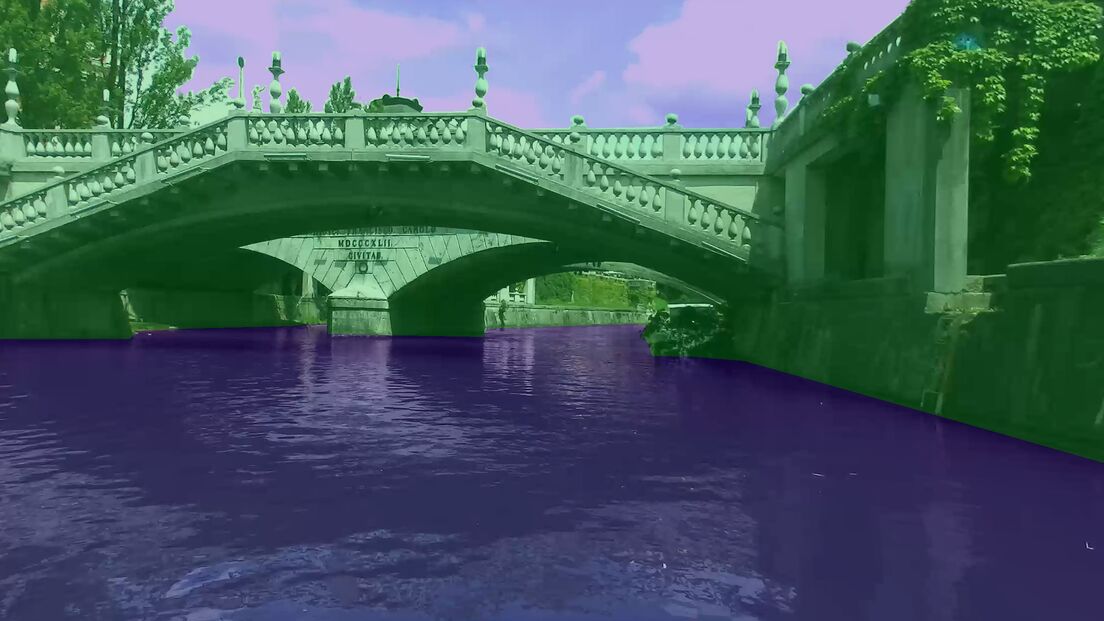}}\\

    \vspace{0.25cm}

  \end{tabular}
  \egroup
  
  \caption{Examples of data from our dataset and corresponding semantic labels. The upper right image depicts LIDAR points overlaid on the RGB image, while the bottom row shows thermal camera image and semantic annotations overlays, respectively. The semantic labels of \textit{sky}, \textit{water} and \textit{static obstacle} are denoted with purple, blue, and green respectively.}
  \label{fig:title}
  
\end{figure*}

While adding additional sensors increases the complexity of the sensor system, extra modalities can be invaluable for accurate scene interpretation, especially under aforementioned difficult circumstances. Different modalities can provide various benefits for the reliability of the system. While color images are easy to interpret under normal circumstances and carry a lot of information, specific parts of the scene can be discerned more easily using different sensors. A polarization camera can be used to distinguish between reflections and objects due to multiple polarization angles, while objects on water can frequently be spotted easily using a thermal camera because of the temperature differences and differences between materials' emissivity. Additionally, LIDAR does not detect the water surface, thus serving as an additional source of information regarding the areas that are safe for navigation or as a verification of the presence of obstacles on the water surface.

Using many sensors is far from trivial, since both temporal and spatial alignment are needed to allow for joint interpretation of heterogeneous data streams. Supervised deep learning methods also require large amounts of annotated data, which can be difficult to obtain due to non-RGB data being hard to interpret for human annotators. In this paper, we present a maritime dataset containing multiple modalities and introduce a method for training robust multimodal architectures while only training using RGB annotations. We also report on our approach for aligning and annotating the data captured using our system. An example of our aligned and annotated data is shown in Figure~\ref{fig:title}.

\subsection{Contributions}

We present a novel multimodal dataset MULTIAQUA (Multimodal Aquatic Dataset), containing five different image modalities, as well as point cloud data from LIDAR and RADAR sensors along with GPS and IMU measurements. Six of the system modalities, namely the cameras and LIDAR were precisely calibrated using the method from~\cite{muhovivc2023joint}, thus all their data can be spatially aligned. The remaining sensors' extrinsic parameters were determined manually. We provide pixel-level annotations for RGB images, as well as the methods for transferring the annotations to other sensors or vice versa. Fully supervised multimodal semantic segmentation methods can therefore be applied to the data. Additionally, we present a training approach for existing multimodal architectures that enables better exploitation of different modalities. Our proposed changes drastically improve the models' performance under difficult circumstances, such as poor lighting conditions, despite only training the models on daytime data. The improvements in performance are demonstrated with extensive experimental work on our difficult test set and also other multimodal datasets.


\section{Related work}

Along with the field of autonomous ground vehicles, the field of USV navigation has been flourishing in recent years. Large and detailed datasets for automotive industry~\cite{cordts2016cityscapes, sun2020scalability, geiger2013vision, maddern2017} and strong general methods such as DeepLab~\cite{chen2017deeplab}, SegFormer~\cite{xie2021segformer}, and YOLO~\cite{wang2023yolov7, yolov8_ultralytics}, laid the foundation for maritime-specific methods and datasets. These are required, since the methods and datasets devised for the AGV domain often fail to perform well enough for the marine domain. Simple obstacle detection methods based on hand-crafted features~\cite{Cho2015AutonomousDA}, statistics~\cite{kristan2015fast}, saliency estimation~\cite{cane2016saliency} and background subtraction~\cite{prasad2018object} were first applied to the maritime domain. Conventional CNN-based methods~\cite{lee2018image, yang2019surface, bovcon2021mods} were later used in the maritime domain as well. While including many maritime-relevant classes and generally performing well, these methods were still limited to known object classes and not particularly resilient with regard to sun glitter and reflections.

Further research shifted towards semantic segmentation~\cite{cane2018evaluating, chen2021wodis, yao2021shorelinenet} in order to avoid explicitly modeling obstacle classes. In their paper, Bovcon and Kristan~\cite{bovcon2021wasr} presented WaSR, an encoder-decoder based semantic segmentation method that focuses on the separation of semantic features along with using IMU data to predict pixel labels \textit{sky}, \textit{water} and \textit{obstacle}. The method was extended by Žust et al.~\cite{vzust2022temporal} to improve performance by using temporal information and reducing false positives caused by the water surface.
With the rise of transformer-based approaches, general semantic segmentation methods appear to have reached a level that can outperform maritime-specific methods. The analysis by Žust et al.~\cite{zust2023lars} shows that general single-image state-of-the art methods can significantly outperform maritime-specific and temporal semantic segmentation methods --- the highest performing methods in this case being KNet~\cite{zhang2021k}, SegFormer~\cite{xie2021segformer} and DeepLabv3+~\cite{chen2018encoder}.

\subsection{Datasets}

Many automotive datasets have been published in the past to allow for faster development of autonomous ground vehicles, such as KITTI~\cite{geiger2013vision}, Waymo~\cite{sun2020scalability}, Cityscapes~\cite{cordts2016cityscapes}, nuScenes~\cite{caesar2020nuscenes}, and Oxford RobotCar~\cite{maddern2017}. These datasets usually include several RGB cameras, LIDAR and GPS/IMU. Some datasets also include stereo cameras to provide depth information. This set of sensors is rarely extended with other sensors, with one exception being the DENSE dataset~\cite{bijelic2020seeing} that also includes radar and gated NIR data for the purpose of navigating adverse conditions. Aside from automotive research, several multimodal datasets have been proposed for different tasks such as indoor semantic segmentation(NYUv2~\cite{silberman2012indoor}), pedestrian detection at night (MFNet~\cite{ha2017mfnet}, KAIST~\cite{hwang2015multispectral}, LLVIP~\cite{jia2021llvip}), material segmentation (MCubeS~\cite{Liang_2022_CVPR}), SLAM (ViViD~\cite{lee2019vivid}, ViViD++~\cite{lee2022vivid++}), and image restoration (MOFA~\cite{xiao2023mofa}). Synthetic datasets including many different modalities have also been designed, notably the recent DeLiVER~\cite{zhang2023delivering}, SHIFT~\cite{sun2022shift}, and SELMA~\cite{testolina2023selma} datasets which include many modalities, such as RGB, depth, LIDAR and event camera, as well as precise ground truth annotations and diverse scenes.

Large comprehensive datasets in the maritime domain are comparatively rare. Some were presented in the past, such as SMD~\cite{smd_prasad2017video}, MODD~\cite{kristan2015fast} and MODD2~\cite{bovcon2018stereo}, which contain obstacle annotations using bounding boxes but are fairly conservative in diversity and size. A training semantic segmentation dataset, MaSTr1325~\cite{bovcon2019mastr}, was also released, containing per-pixel annotations for classes \textit{obstacle}, \textit{water} and \textit{sky}. Several other datasets followed suit, for example ROSEBUD~\cite{lambert2022rosebud}, Waterline~\cite{steccanella2020waterline} and Tampere-WaterSeg~\cite{taipalmaa2019high}. These datasets all extend the diversity of maritime annotations, containing images captured on rivers and lakes around the world, but mostly do not contain visually very challenging scenarios. Bovcon et al.~\cite{bovcon2021mods} recently presented MODS, a benchmark dataset that addresses both obstacle detection and semantic segmentation evaluation. The issue of low data diversity and scale was recently addressed with LaRS~\cite{zust2023lars}, which boasts a large number of fully panoptically annotated maritime images. The dataset covers a very large spectrum of difficult visual conditions that appear in marine environments, such as different times of day, reflections, water surface appearance and adverse weather conditions. This is so far the most comprehensive monocular marine dataset in existence, and its purpose is to facilitate learning of models for panoptic segmentation.

\begin{table*}[h]
    \centering
    \caption{Overview of maritime datasets, their available sensor modalities, and provided annotations. The presence of a modality in a dataset is marked with a checkmark (✔), while its absence is marked with a cross (✗). \textbf{NIR} - Near infrared camera, \textbf{TIR} - Thermal infrared camera. Annotations: \textbf{BB} – Bounding Boxes, \textbf{SS} – Semantic Segmentation, \textbf{PS} – Panoptic Segmentation, \textbf{WEA} – Water Edge Annotations, }
    \renewcommand{\arraystretch}{1.3}
    \resizebox{\textwidth}{!}{ 
    \begin{tabular}{|l|c|c|c|c|c|c|c|c|c|}
        \hline
        \textbf{Dataset} & \textbf{RGB} & \textbf{LIDAR} & \textbf{Radar} & \textbf{NIR} & \textbf{Echosounder} & \textbf{TIR} & \textbf{Polarization} & \textbf{GPS} & \textbf{Annotations} \\
        \hline
        SMD~\cite{smd_prasad2017video} & ✔ & ✗ & ✗ & ✗ & ✗ & ✗ & ✗ & ✗ & BB \\
        MODD~\cite{kristan2015fast} & ✔ & ✗ & ✗ & ✗ & ✗ & ✗ & ✗ & ✗ & BB \\
        MODD2~\cite{bovcon2018stereo} & ✔ & ✗ & ✗ & ✗ & ✗ & ✗ & ✗ & ✗ & BB \\
        MaSTr1325~\cite{bovcon2019mastr} & ✔ & ✗ & ✗ & ✗ & ✗ & ✗ & ✗ & ✗ & SS \\
        ROSEBUD~\cite{lambert2022rosebud} & ✔ & ✗ & ✗ & ✗ & ✗ & ✗ & ✗ & ✗ & SS \\
        Waterline~\cite{steccanella2020waterline} & ✔ & ✗ & ✗ & ✗ & ✗ & ✗ & ✗ & ✗ & SS \\
        Tampere-WaterSeg~\cite{taipalmaa2019high} & ✔ & ✗ & ✗ & ✗ & ✗ & ✗ & ✗ & ✗ & SS \\
        MODS~\cite{bovcon2021mods} & ✔ & ✗ & ✗ & ✗ & ✗ & ✗ & ✗ & ✗ & BB, SS \\
        LaRS~\cite{zust2023lars} & ✔ & ✗ & ✗ & ✗ & ✗ & ✗ & ✗ & ✗ & PS \\
        ROAM@CRAS~\cite{campos2022modular} & ✔ & ✔ & ✗ & ✗ & ✔ & ✗ & ✗ & ✔ & ✗ \\
        USVInland~\cite{cheng2021we} & ✔ & ✔ & ✔ & ✗ & ✗ & ✗ & ✗ & ✔ & WEA \\
        Flow~\cite{cheng2021flow} & ✔ & ✗ & ✔ & ✗ & ✗ & ✗ & ✗ & ✗ & BB \\
        MassMIND~\cite{nirgudkar2023massmind} & ✗ & ✗ & ✗ & ✗ & ✗ & ✔ & ✗ & ✗ & SS \\
        WaterScenes~\cite{yao2024waterscenes} & ✔ & ✗ & ✔ & ✗ & ✗ & ✗ & ✗ & ✗ & BB, SS, PS \\
        OASIs~\cite{kim2024introducing} & ✔ & ✗ & ✗ & ✗ & ✗ & ✗ & ✗ & ✗ & SS \\
        Polaris~\cite{choi2024polaris} & ✔ & ✔ & ✔ & ✗ & ✗ & ✔ & ✗ & ✗ & BB \\
        MULTIAQUA (Ours) & ✔ & ✔ & ✔ & ✔ & ✗ & ✔ & ✔ & ✔ & SS \\
        \hline
    \end{tabular}
    } 
    \label{tab:maritime_datasets}
\end{table*}

Compared to datasets containing only color images, whether in mono or stereo setups, multimodal maritime datasets are few. In recent years, more additional modalities are being used on waterborne vehicles to include richer information about the observed scenes. Campos et al.~\cite{campos2022modular} presented the ROAM@CRAS dataset, including cameras, a multibeam echosounder, and LIDAR. Cheng et al.~\cite{cheng2021we} presented the USVInland dataset, including both LIDAR and radar data, as well as water edge annotations that can be used for supervised learning. A separate dataset named Flow was also presented by Cheng et al.~\cite{cheng2021flow}, including color images coupled with radar, including annotations of waterborne debris. Nirgudkar et al.~\cite{nirgudkar2023massmind} presented MassMIND, a LWIR dataset that can be used to train semantic segmentation models robust to sun glare and low-light scenarios. Recently Yao et al.~\cite{yao2024waterscenes} presented WaterScenes, a multi-task radar-camera fusion dataset, as well as the corresponding benchmarks for object detection, semantic segmentation and panoptic segmentation. An overview of maritime-specific datasets is depicted in Table~\ref{tab:maritime_datasets}.

The variability of scenes and modalities included in maritime datasets is increasing lately, with more datasets including varied lighting and weather conditions, including sunset and nighttime (c.f. LaRS~\cite{zust2023lars}, WaterScenes~\cite{yao2024waterscenes}, OASIs~\cite{kim2024introducing}). However, there are no maritime datasets thus far that include both very difficult low-light scenarios alongside sensors that could potentially be useful for navigating under such conditions. Adding more sensors increases the power consumption, weight and system complexity, thus increasing the cost of operation and annotation for such systems. Additionally, capturing maritime data in low-light or poor weather circumstances can be difficult and dangerous, thus slowing the acquisition of such samples. With MULTIAQUA, we aim to provide the first dataset that can be used for developing more robust methods that will be able to function even in very difficult circumstances by virtue of efficient exploitation of different modalities.

\subsection{Multimodal methods}

Auxiliary (non-RGB) modalities can be used to enhance the performance of visual methods when only RGB data is not sufficient. The methods that use multiple streams of aligned data use various approaches to data fusion. The creation of highly informational synthetic images has become a research topic on its own\cite{tang2023divfusion, tang2023rethinking}. If data is processed separately, the methods can generally opt for early or late fusion, with more exotic approaches being used in specific domains or architectures. Early fusion (or input fusion) usually denotes approaches where pixel-aligned data is merged into a single tensor and processed like a regular image. This allows well-researched image processing neural networks to jointly learn feature extractors for these synthetic input images. Late fusion methods separately process modalities and merge the results just before producing the final predictions. In multi-stage architectures, intermediate feature fusion is also possible, fusing and combining features from different modalities at each stage of the architecture. 

Recently, several multimodal deep learning architectures for semantic segmentation have been proposed. The most prevalent variation of such methods uses RGB images along with an extra auxiliary modality, such as MFNet~\cite{ha2017mfnet} and CMX~\cite{zhang2023cmx}. Including more than two modalities is fairly rare because it can significantly increase the architecture complexity and memory requirements. However, recently Zhang et al.~\cite{zhang2023delivering} proposed a multimodal method CMNeXt that can support an arbitrary number of auxiliary modalities with minimal increase in architecture complexity and memory footprint. The authors show increased performance on different multimodal datasets and even proposed their own synthetic dataset DeLiVER that includes RGB images, depth data, LIDAR and event data. Their approach is based on Segformer~\cite{xie2021segformer} backbone for extracting features from modalities, but only includes two copies of the backbone regardless of the number of additional modalities. The selection of informative features from additional modalities is performed by a Self-Query Hub, and the additional modalities are fused using the feature rectification and fusion modules presented in CMX architecture~\cite{zhang2023cmx}. Working in a similar manner, the MMSFormer architecture proposed by Reza et al.~\cite{reza2024mmsformer} uses modality-specific encoders to process different modalities, then uses a fusion block to process them and produce the final multimodal features that can finally be used for inferring the semantic segmentation. The authors claimed the best performance on MCubeS~\cite{Liang_2022_CVPR}, FMB~\cite{liu2023multi}, and PST900~\cite{shivakumar2020pst900} datasets.

Recently, Li et al.~\cite{li2024stitchfusion} proposed StitchFusion, a versatile multimodal method that introduces the MultiAdapter Layer as an information exchange mechanism between modalities. The module consists of downscaling, processing by a non-linear function, then upscaling back to the original dimension. In their architecture, all pairs of used modalities use an independent bidirectional MultiAdapter, for optimal exploitation of cross-modality information. The authors achieved the best performance on MCubeS~\cite{Liang_2022_CVPR}, DeLiVER~\cite{zhang2023delivering}, FMB~\cite{liu2023multi} and MFNet~\cite{ha2017mfnet} datasets.

\subsection{Nighttime domain adaptation}
Several approaches for handling difficult weather and lighting circumstances have been proposed, mostly in the domain of autonomous ground vehicles~\cite{zhang2023complementary}. Vertens et al.~\cite{vertens2020heatnet} proposed HeatNet for training RGB and thermal image semantic segmentation on both daytime and nighttime data. They explored the teacher-student approach, in which pretrained segmentation models (trained separately on RGB and thermal images) guide the training process for the multimodal architecture. They also presented an RGB-thermal dataset named Freiburg Thermal. Similarly, DANNet~\cite{wu2021dannet} was proposed, which tries to align features from daytime and nighttime images to improve on semantic segmentation. This is enabled by the Dark Zurich dataset, which includes daytime and nighttime images along with corresponding GPS data. Thus, spatially aligned images captured under different illumination conditions can be acquired and a relighting network can be applied to bring image features closer together and allow for better segmentation. Romera et al.~\cite{romera2019bridging} presented a GAN-based approach for transforming nighttime images to improve the semantic segmentation performance. Xia et al.~\cite{xia2023cmda} developed an approach to fusing color images and event camera data to address the domain adaptation problem. Two datasets are commonly used for training and evaluating new nighttime driving approaches: Nighttime driving~\cite{dai2018dark} and Dark Zurich~\cite{sakaridis2019zurich}.

\section{MULTIAQUA sensor system}
\label{sec:sensors}

The sensor system used in the acquisition of our proposed dataset consists of several modules, each being responsible for one sensor. Each module includes an embedded computer that manages the sensor initialization, local data storage and communication. The communication is implemented over UDP protocol and allows for real-time simultaneous streaming of data during recording sessions. The sensor data is stored locally without the need for network transfer. The computers are synchronized using NTP to ensure accurate timestamps for data packets and images. The sensor selection was focused on exploiting phenomena that could help in scene interpretation and navigation under difficult environmental conditions. Specifically, the distinction between water and non-water parts of the environment and detection of floating obstacles were the main goals of the sensor system setup.

\noindent The sensors included in our system are as follows:
\begin{itemize}
\itemsep0.3em
    \item Velodyne VLP-16 LIDAR
    \item Stereolabs ZED 2 (RGB stereo camera)
    \item Dual-spectrum NIR camera (RaspiCam NoIR with filters)
    \item Teledyne FLIR polarization camera (BFS-U3-51S5P-C)
    \item Thermographic (LWIR) Device-ALab camera (SmartIR384L).
    \item Smartmicro automotive radar (UMRR-11 Type 132)
    \item GPS/IMU module with real-time DGPS correction (Advanced Navigation Spatial)
\end{itemize}

\section{Dataset}

Here we present our multimodal dataset, named MULTIAQUA (Multimodal Aquatic Dataset). The dataset was captured using the multimodal sensor stack described in Section~\ref{sec:sensors}. The dataset contains 3293 frames, sampled from 6 separate recording sessions. Synchronized data from all sensors is provided for each of the included frames, if available (c.f. Table~\ref{tab:dataset}). We publish all the acquired sensor data for the selected frames, regardless of perceived quality. While some nighttime images are very dark and noisy, there might be some information present that is not observable by the naked eye. The exception is the NIR data from sequence \textit{lj4} which is nearly fully black, and we omitted these images deliberately. Manual per-pixel semantic labels are available for 3093 of the included RGB images. The remaining 200 frames have the corresponding semantic labels based on the thermal camera images (more detail on the annotation scheme is provided in Section~\ref{sec:annotation}).

\begin{figure}[ht!]
    \centering
    \def\twidth{0.3} 

    \bgroup
    \def\arraystretch{1.0}
    
    \begin{tabular}{>{\centering\arraybackslash}m{\twidth\columnwidth} 
                    >{\centering\arraybackslash}m{\twidth\columnwidth} 
                    >{\centering\arraybackslash}m{\twidth\columnwidth}}

    \frame{\includegraphics[width=\twidth\columnwidth]{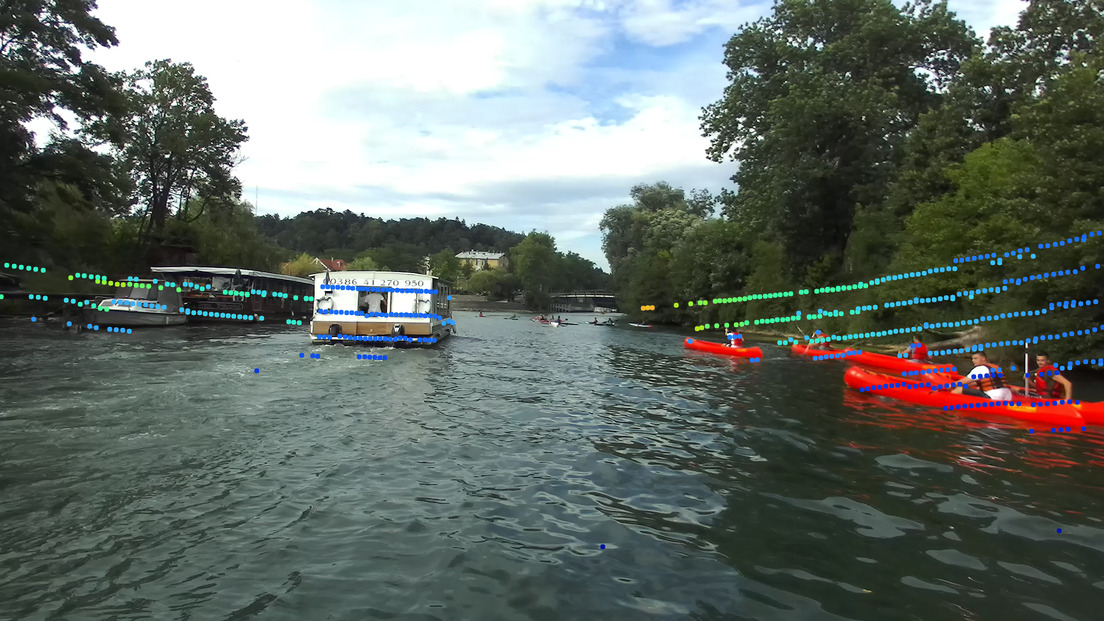}} &
    \frame{\includegraphics[width=\twidth\columnwidth]{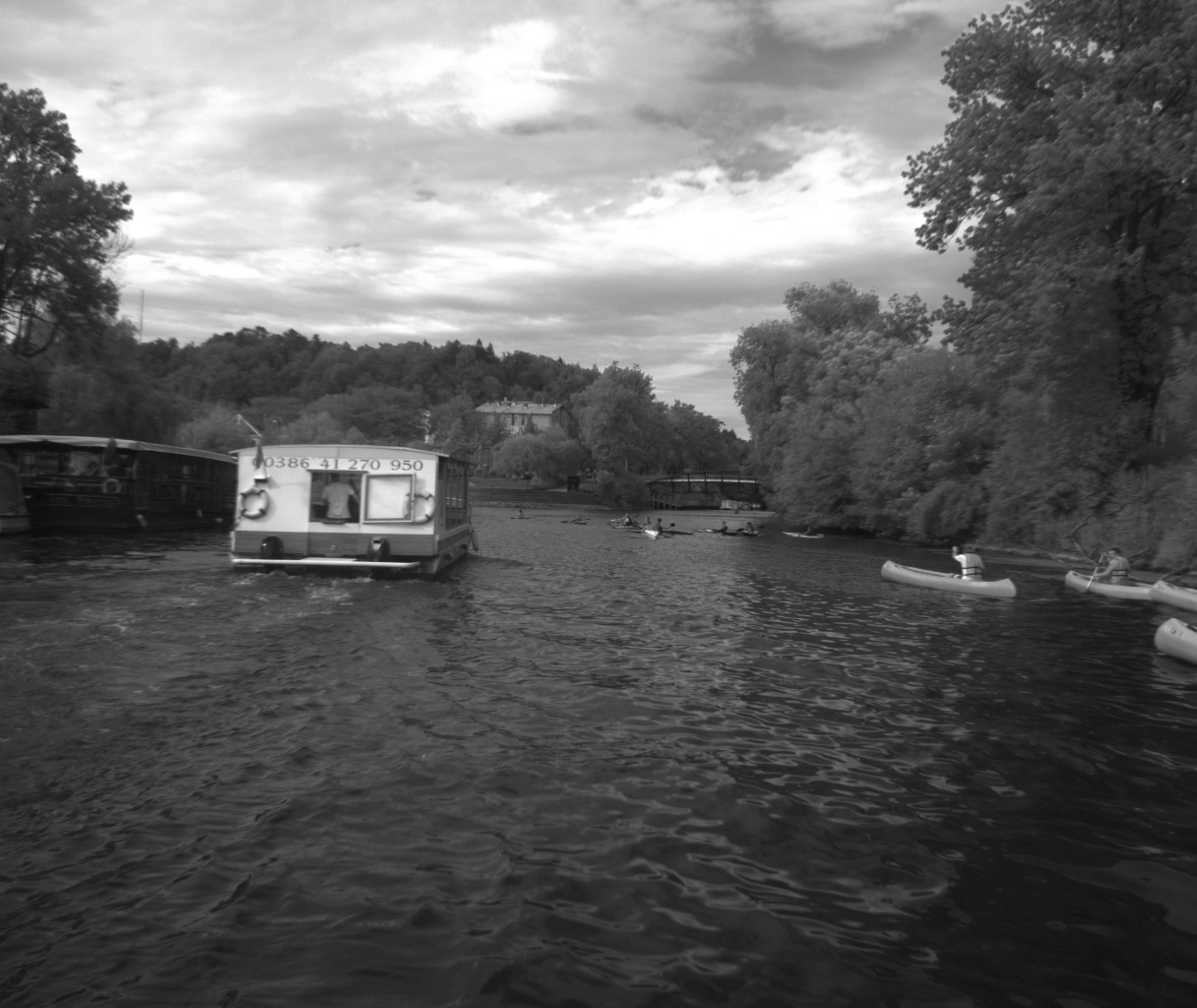}} &
    \frame{\includegraphics[width=\twidth\columnwidth]{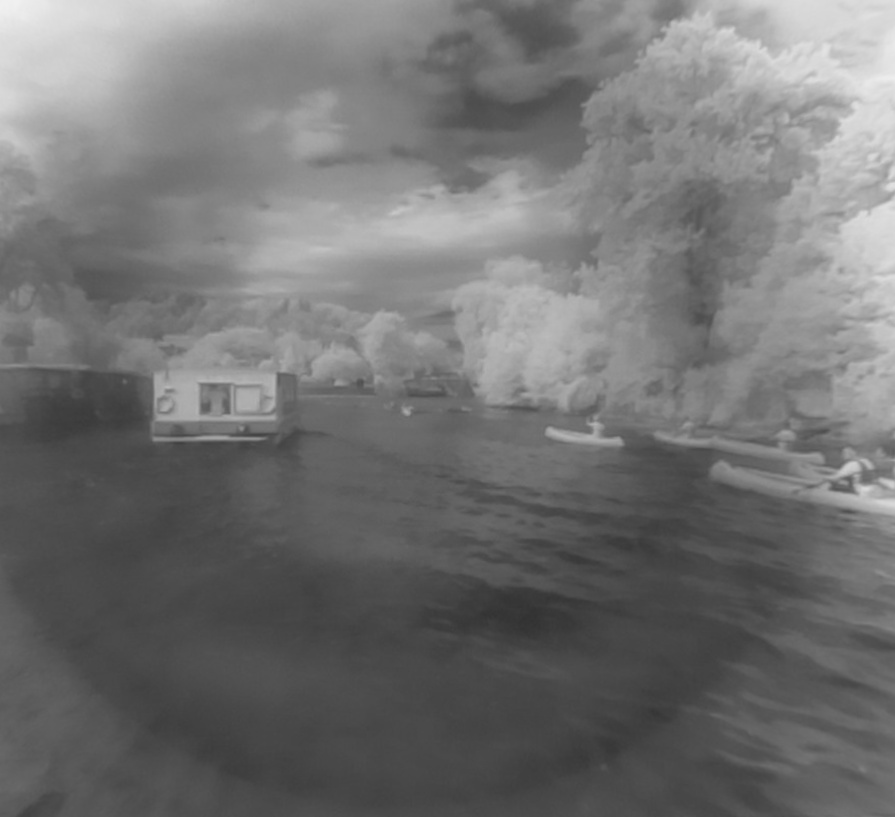}} \\[0.5cm]
    
    ZED + LIDAR & Polarization camera & IR1 \\[0.5cm]

    \frame{\includegraphics[width=\twidth\columnwidth]{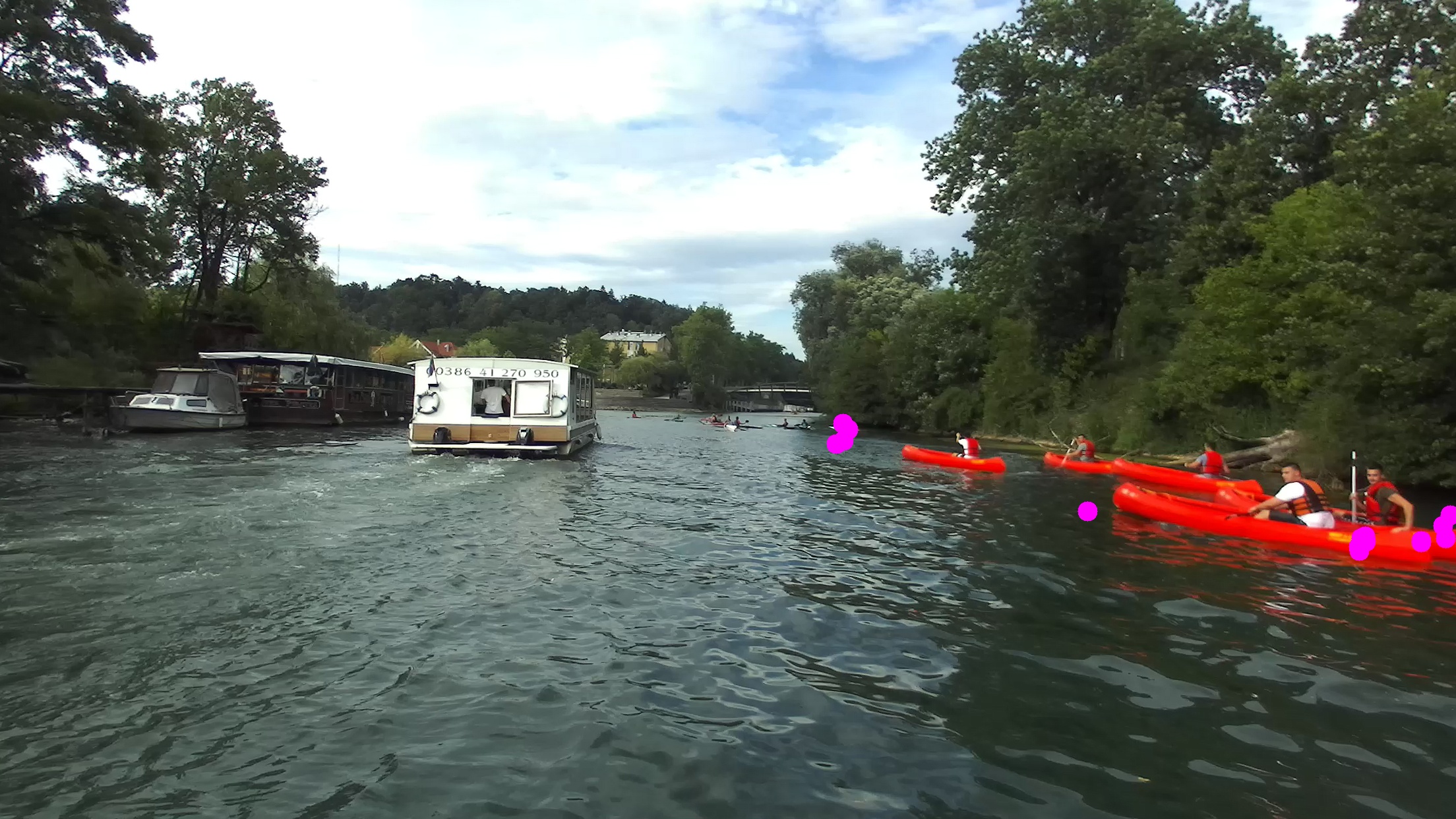}} &
    \frame{\includegraphics[width=\twidth\columnwidth]{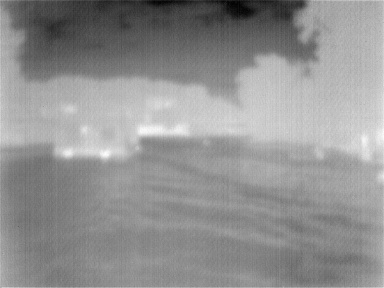}} &
    \frame{\includegraphics[width=\twidth\columnwidth]{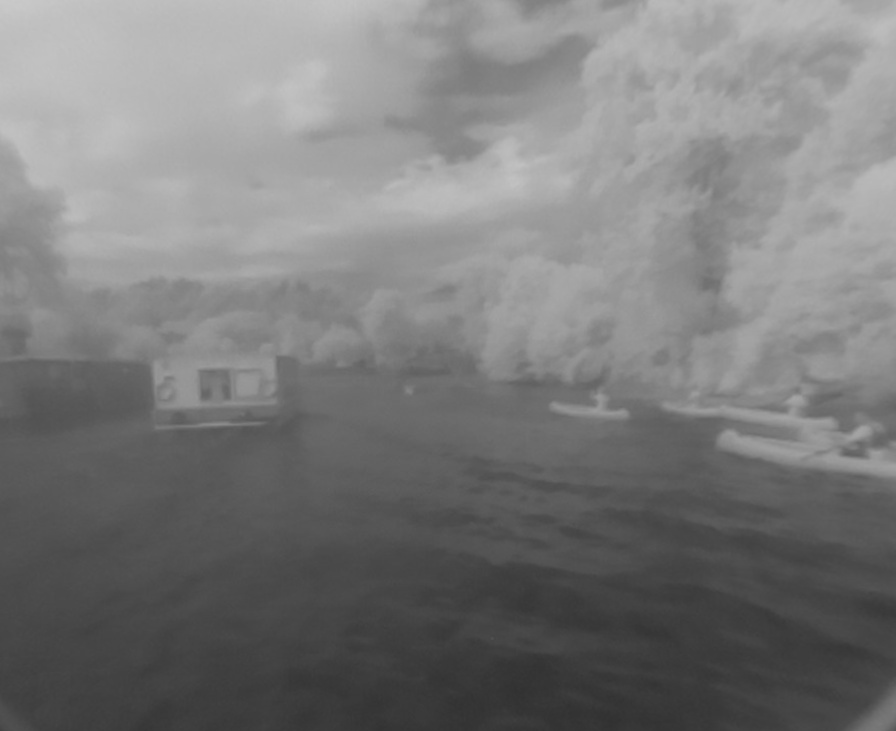}} \\[0.5cm]

    ZED + radar & Thermal camera & IR2 \\[0.5cm]

    \end{tabular}
    \egroup
  
    \caption{Examples of different sensor modalities included in our sensor system. Note the different resolution, focal length, and aspect ratio of the images. The LIDAR points in the upper-left image are colored based on distance. The radar points in the bottom-left image are depicted with purple.}
    \label{fig:dataset_modalities}
  
\end{figure}

Our dataset includes more modalities than any other maritime dataset presented thus far and aims at exploring the usability of sensors before unseen in the maritime domain, such as polarization and thermal cameras. We also developed a calibration approach that aligns the sensors to a common reference and allows for simpler joint annotation of semantic labels for all modalities. This was presented in an earlier work~\cite{muhovivc2023joint}, but is also explained in Section~\ref{sec:structure}. Images of different modalities acquired by our sensor system are depicted in Figure~\ref{fig:dataset_modalities}.

\subsection{Acquisition}
The data included in this dataset was gathered on inland and maritime waterways in Slovenia. The locations recorded are the Ljubljanica river in central Slovenia, lake Bled and the Adriatic coast between Koper and Portorož. The data includes different weather and lighting conditions, and was captured in various parts of the day. Altogether, about 650k frames (or equivalent) were captured on each sensor in about 12 hours of cumulative recording time. The sequences included were each recorded in a single day, and are structured as follows: the dataset consists of 6 sequences, of which four were captured on the Ljubljanica river (\textit{lj1}, \textit{lj2}, \textit{lj3}, and \textit{lj4}), one sequence was captured on lake Bled (\textit{bled1}) and one was captured on the Adriatic coast (\textit{adr1}) 
The weather conditions included in the dataset range from very sunny to overcast, but no precipitation was recorded. The time of day ranges from early morning to early evening, with drastic lighting changes caused by the sun position and consequent high contrast and glitter on the water surface. One of the sequences (\textit{lj4}), however, was captured on the Ljubljanica river at night, including both urban and non-urban areas, with drastic changes in the intensity and structure of illumination.


\begin{table*}[]
\centering
\resizebox{0.8\textwidth}{!}{

\begin{tabular}{c||c|c|c|c|c|c}
\textbf{name} & \textit{lj1} & \textit{lj2} & \textit{lj3} & \textit{lj4} & \textit{bled1} & \textit{adr1} \\ \hline
\textbf{\# frames} & 101666 & 81556 & 118230 & 107890 & 65462 & 139818 \\ \hline
\textbf{\# annotations} & 794 & 598 & 579 & 200 & 344 & 778 \\ \hhline{=#=|=|=|=|=|=}
ZED & ✔ & ✔ & ✔ & ✔ & ✔ & ✔ \\ \hline
polarization & 48\% & 83\% & 69\% & ✔ & 99\% & 98\% \\ \hline
thermal & 97\% & 98\% & 88\% & ✔ & 97\% & 97\% \\ \hline
IR1 & ✗ & ✗ & 97\% & ✗ & ✔ & ✔ \\ \hline
IR2 & ✗ & ✗ & 97\% & ✗ & ✔ & ✔ \\ \hline
radar & ✔ & ✔ & ✔ & ✔ & ✔ & ✔ \\ \hline
LIDAR & ✔ & ✔ & ✔ & ✔ & ✔ & ✔ \\ \hline
GPS & ✔ & ✗ & ✔ & ✔ & 98\% & ✔
\end{tabular}
}
\caption{Presence and coverage of sensors per sequence. Cells are marked with ✔ if data is present for all frames, with ✗ if sensor was unavailable, and with coverage percentage otherwise. The cause of missing data is either technical difficulties during recording, or the data was unusable due to low-light conditions.}
\label{tab:dataset}
\end{table*}

\subsection{Annotation}
\label{sec:annotation}

\begin{figure*}[t!]
    \centering
    \def\twidth{0.23}

    \bgroup
    \def\arraystretch{1.0}
    \begin{tabular}{cccc}

    
    \frame{\includegraphics[width=\twidth\textwidth]{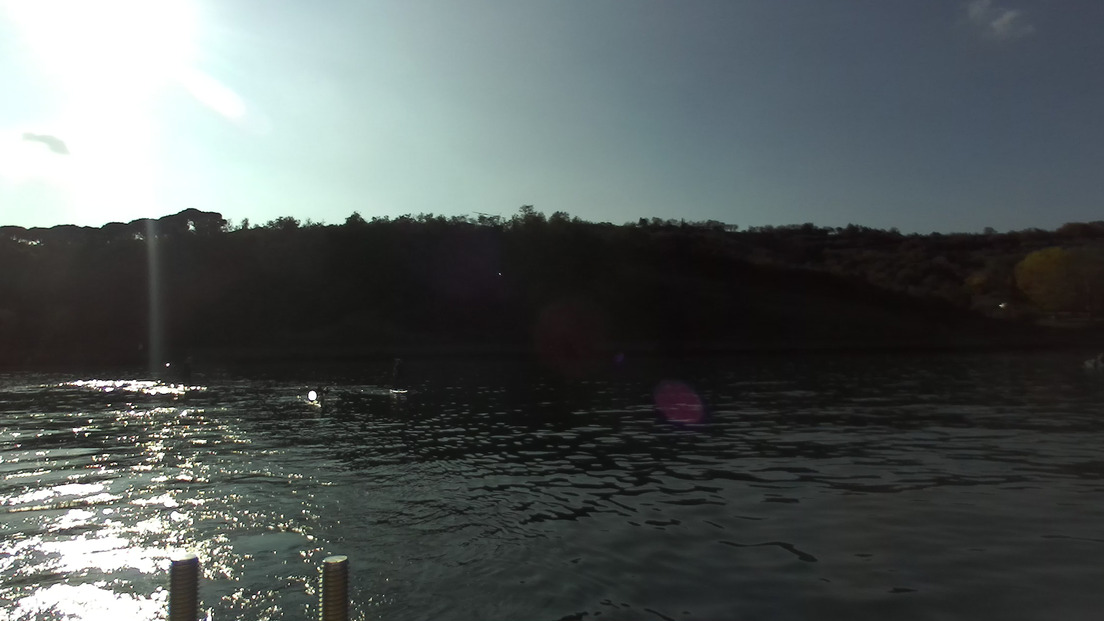}} &
    \frame{\includegraphics[width=\twidth\textwidth]{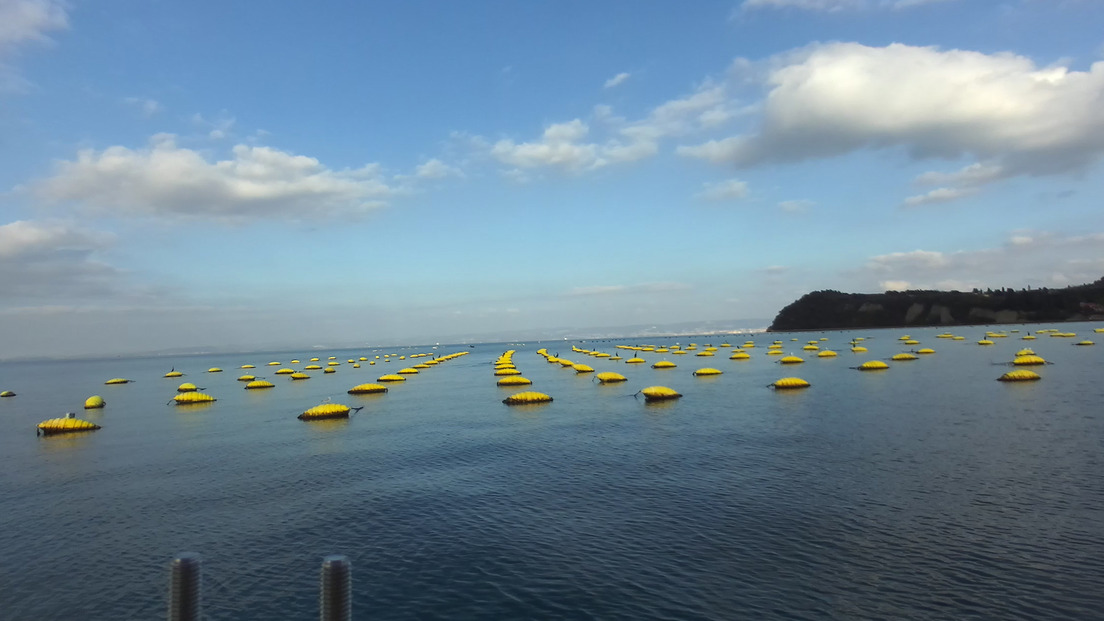}} &
    \frame{\includegraphics[width=\twidth\textwidth]{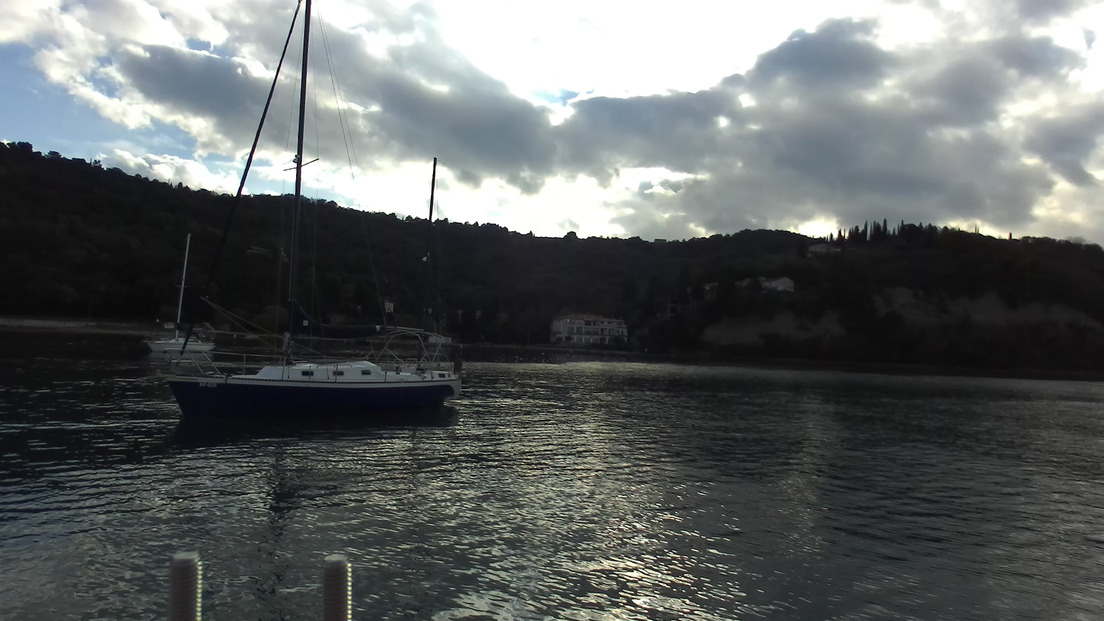}} &
    \frame{\includegraphics[width=\twidth\textwidth]{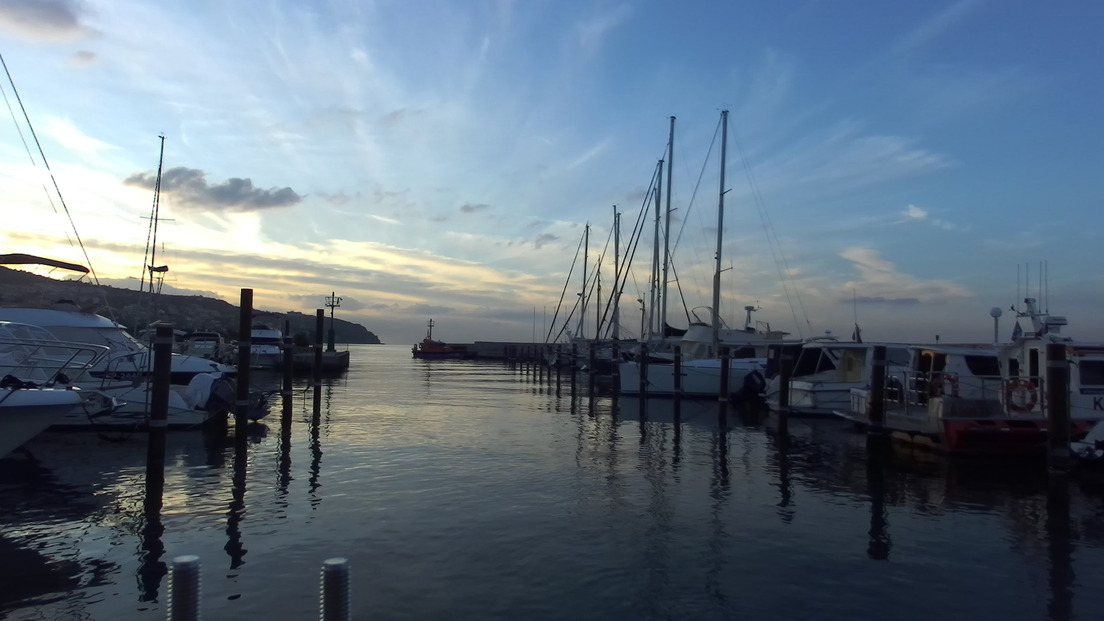}}\\

    \multicolumn{4}{c}{Images captured on the Adriatic seaside (\textit{adr1})} \\[0.5cm]

    \frame{\includegraphics[width=\twidth\textwidth]{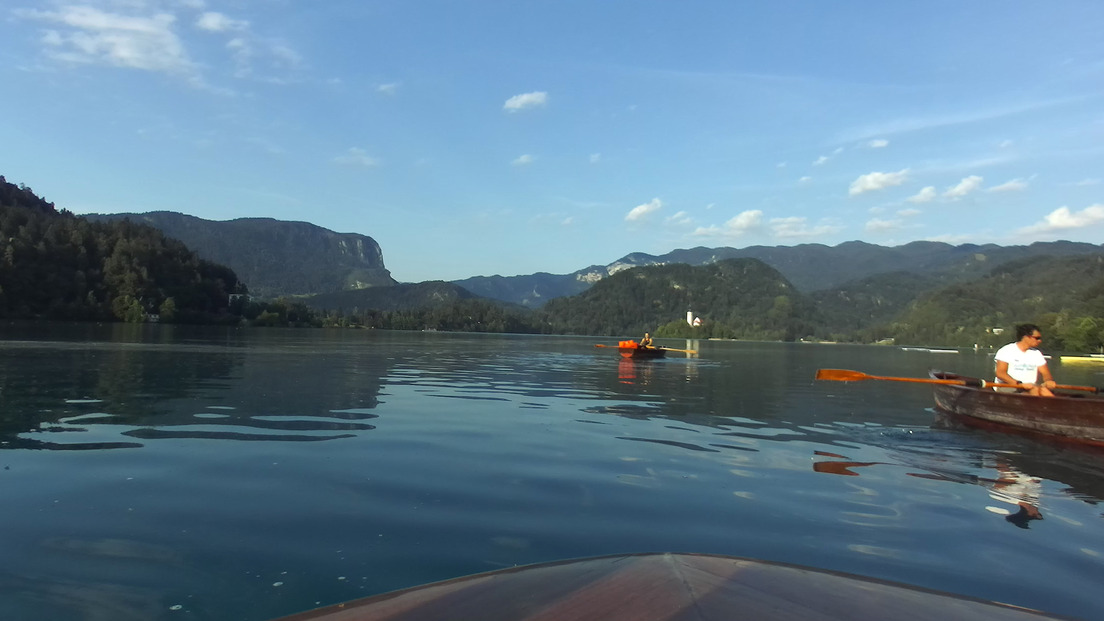}} &
    \frame{\includegraphics[width=\twidth\textwidth]{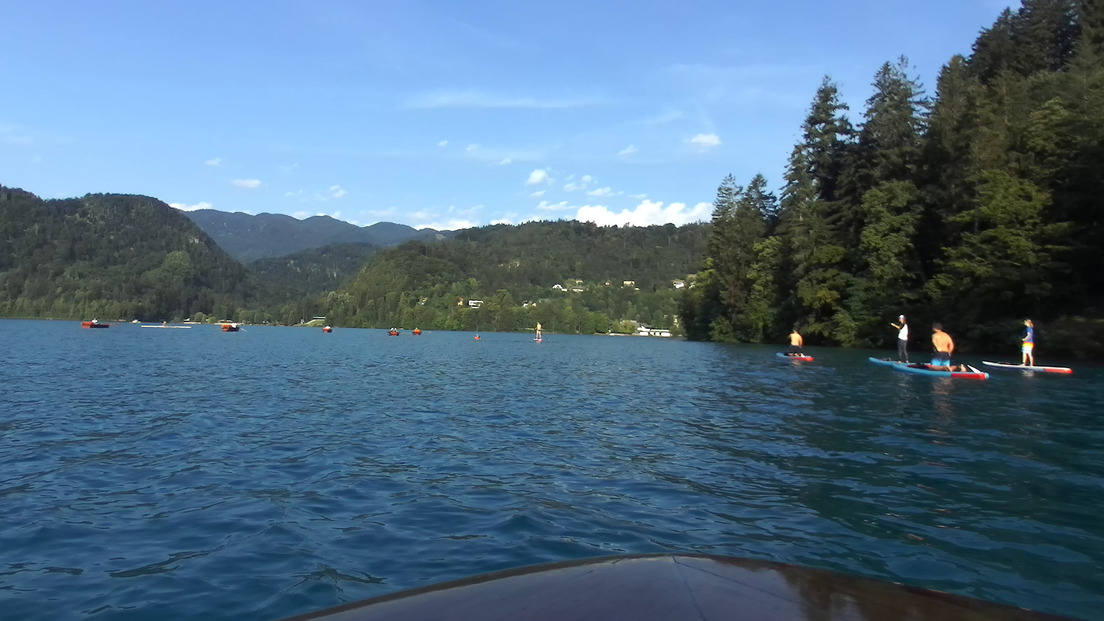}} &
    \frame{\includegraphics[width=\twidth\textwidth]{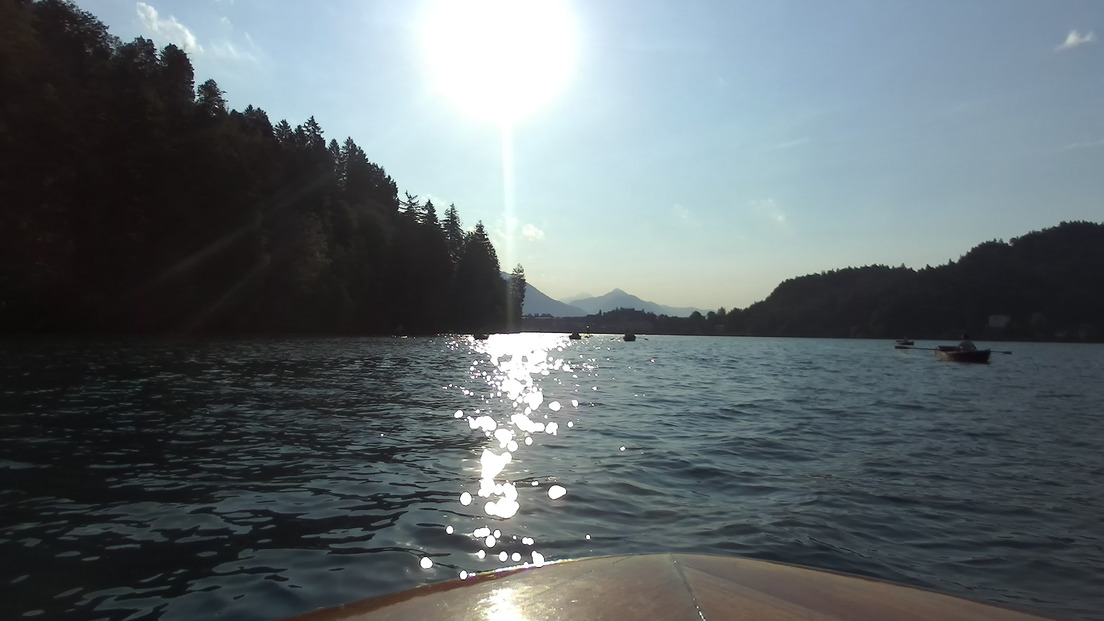}} &
    \frame{\includegraphics[width=\twidth\textwidth]{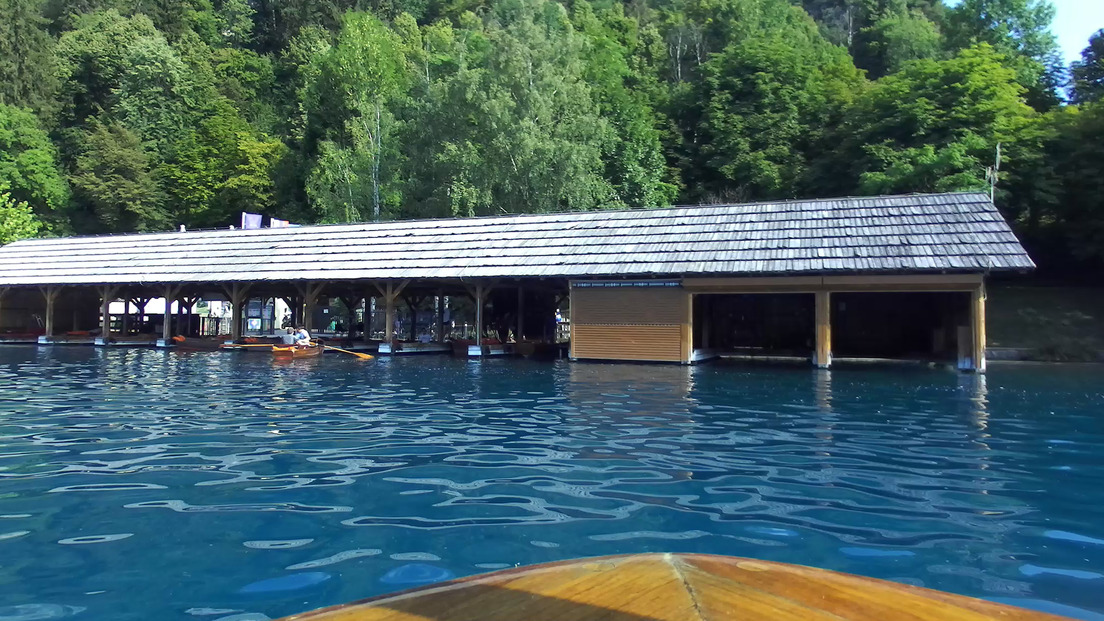}}\\

    \multicolumn{4}{c}{Images captured on Lake Bled (\textit{bl1})} \\[0.5cm]

    \frame{\includegraphics[width=\twidth\textwidth]{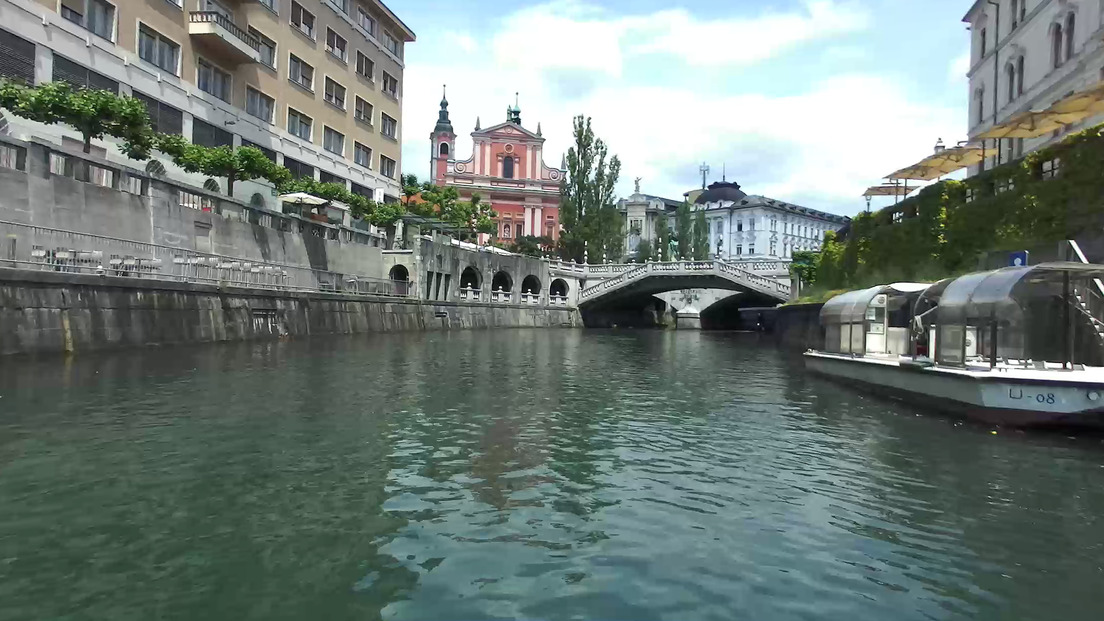}} &
    \frame{\includegraphics[width=\twidth\textwidth]{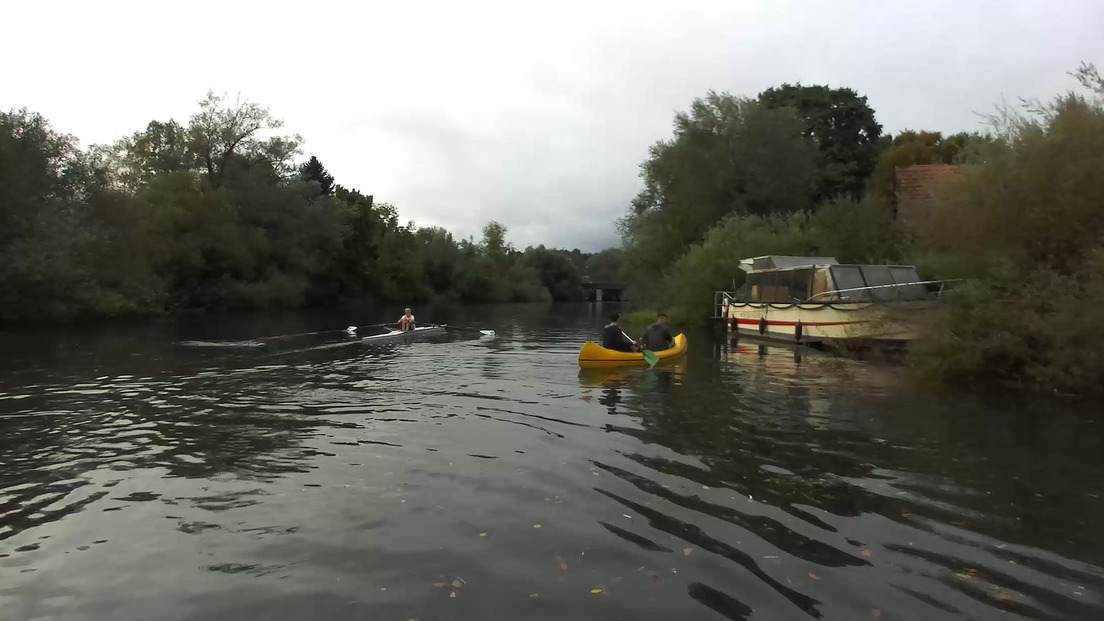}} &
    \frame{\includegraphics[width=\twidth\textwidth]{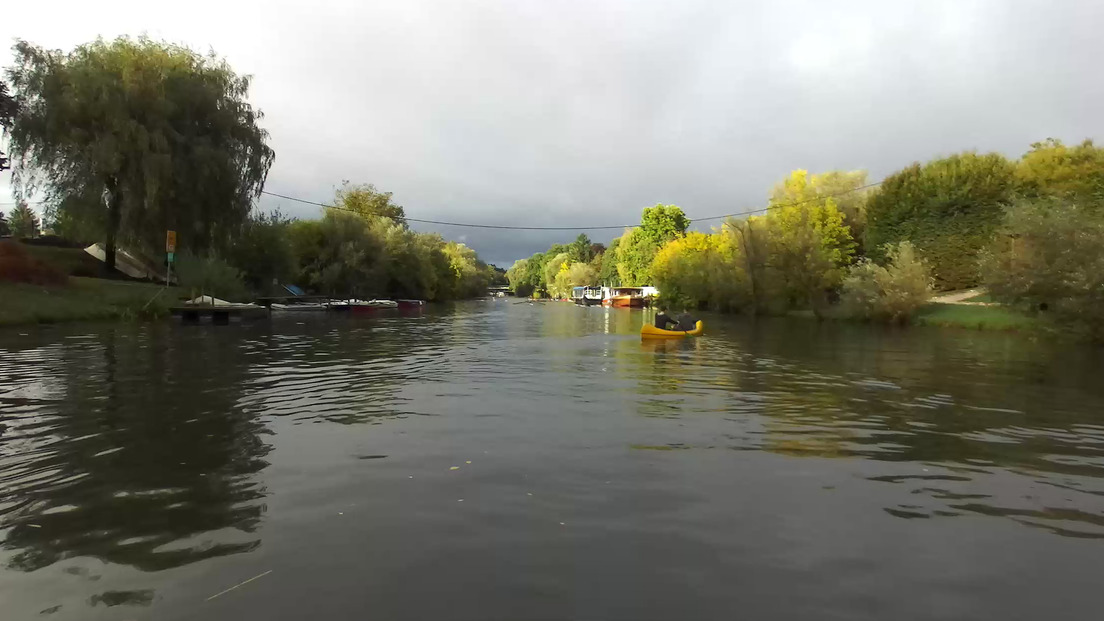}} &
    \frame{\includegraphics[width=\twidth\textwidth]{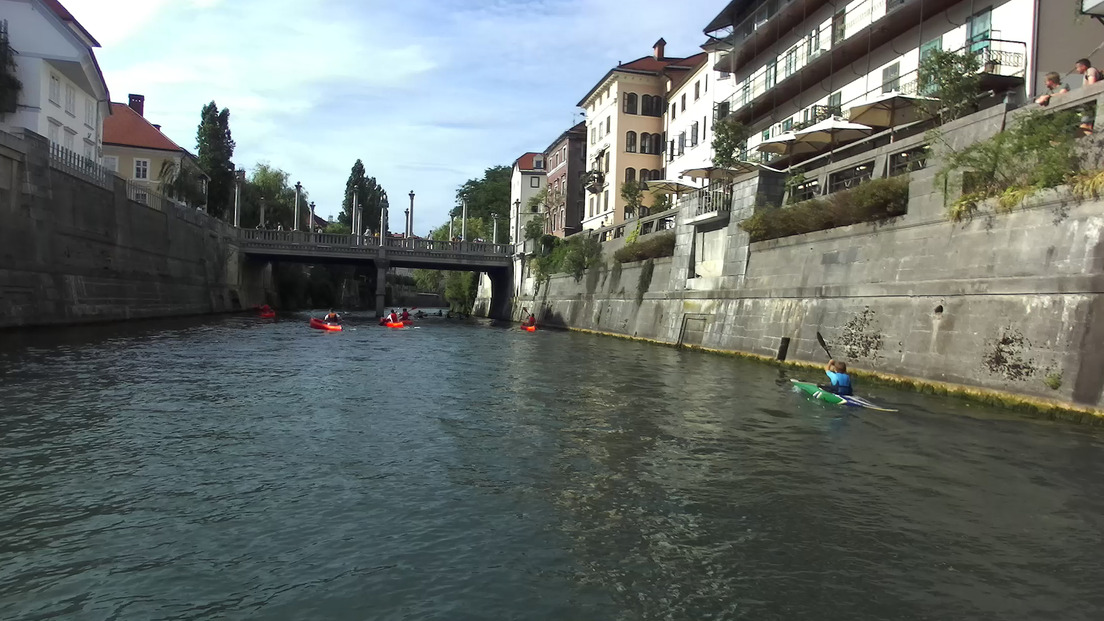}}\\

    \multicolumn{4}{c}{Images captured on the Ljubljanica river during the day (\textit{lj1}, \textit{lj2}, \textit{lj3})} \\[0.5cm]

    \frame{\includegraphics[width=\twidth\textwidth]{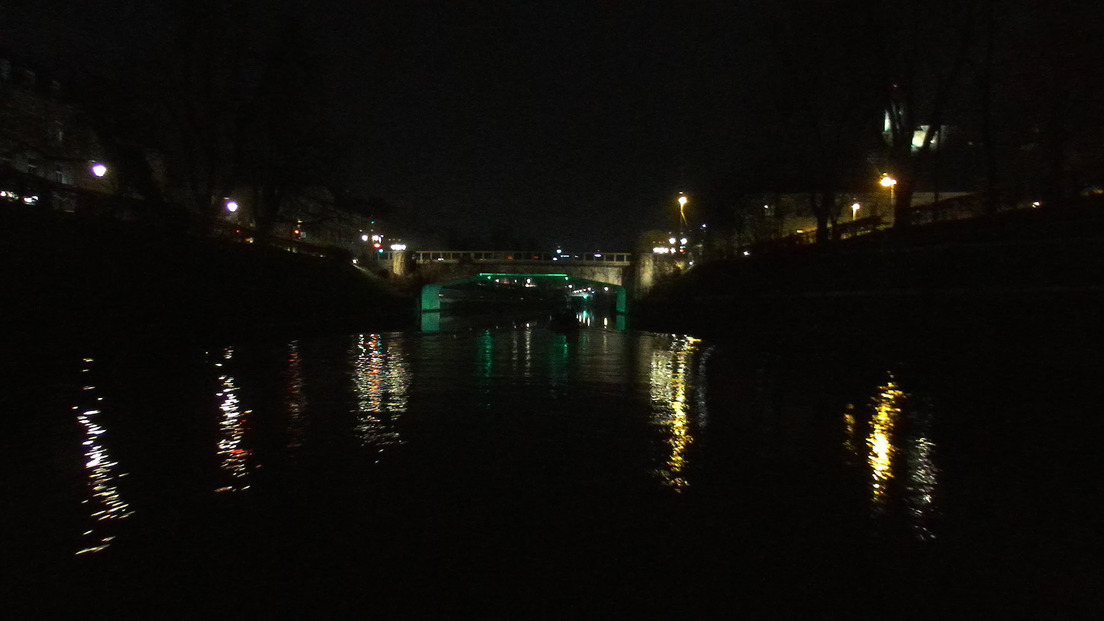}} &
    \frame{\includegraphics[width=\twidth\textwidth]{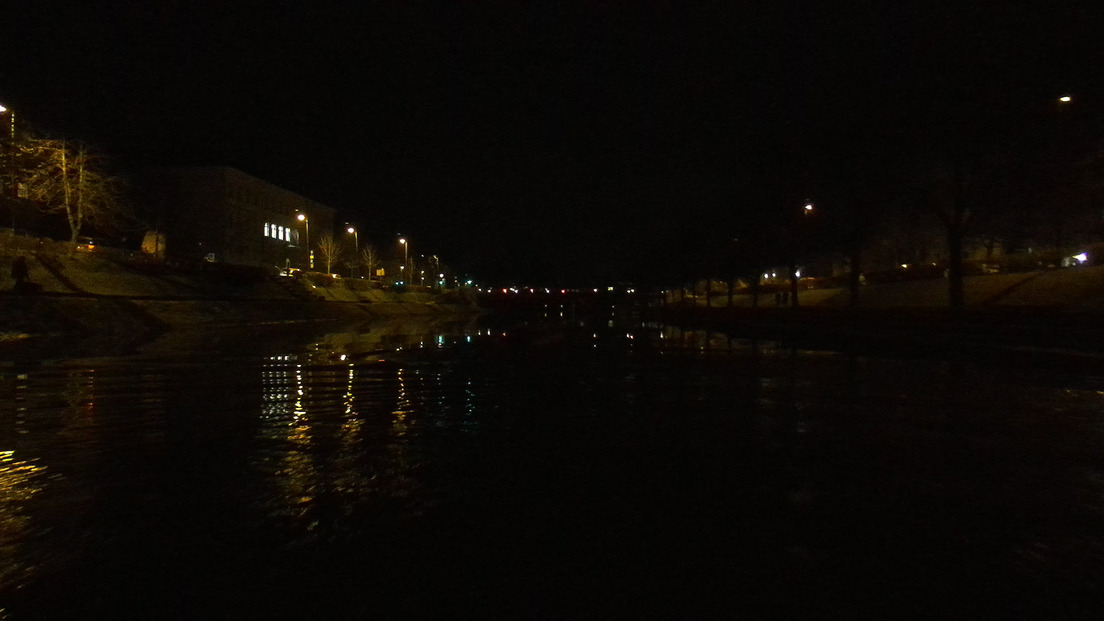}} &
    \frame{\includegraphics[width=\twidth\textwidth]{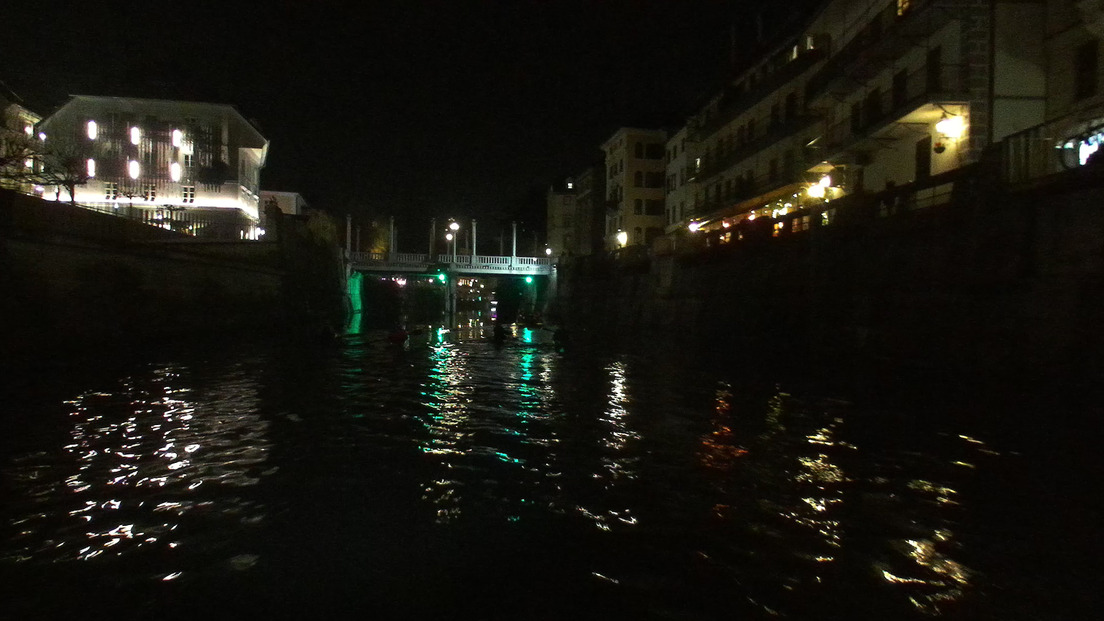}} &
    \frame{\includegraphics[width=\twidth\textwidth]{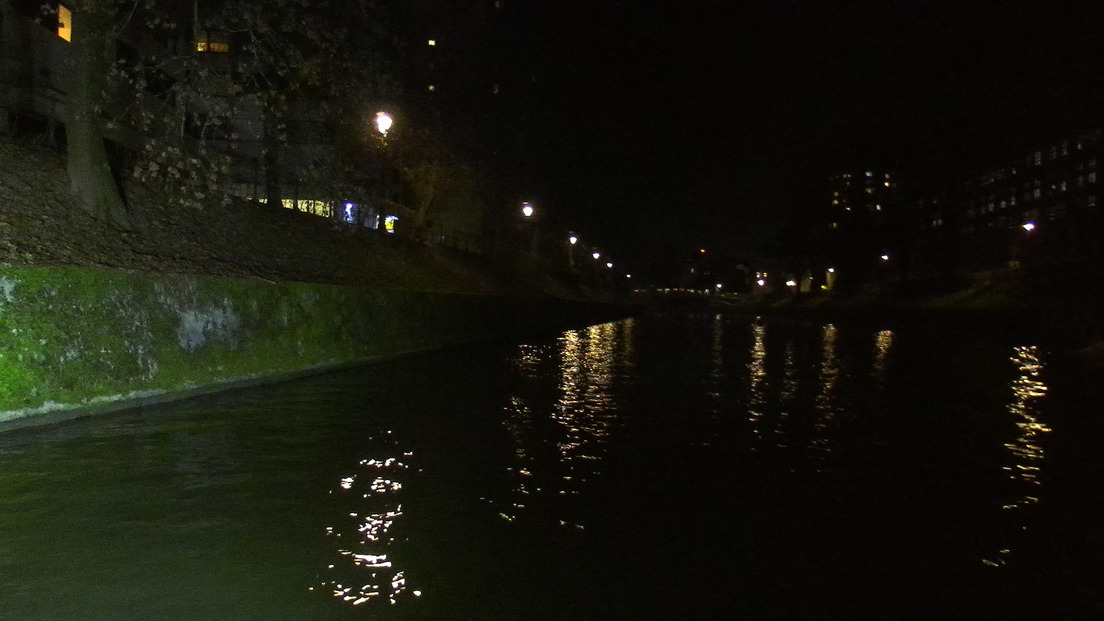}}\\

    \multicolumn{4}{c}{Images captured on the Ljubljanica river at night in urban areas (\textit{lj4})} \\[0.5cm]

    \frame{\includegraphics[width=\twidth\textwidth]{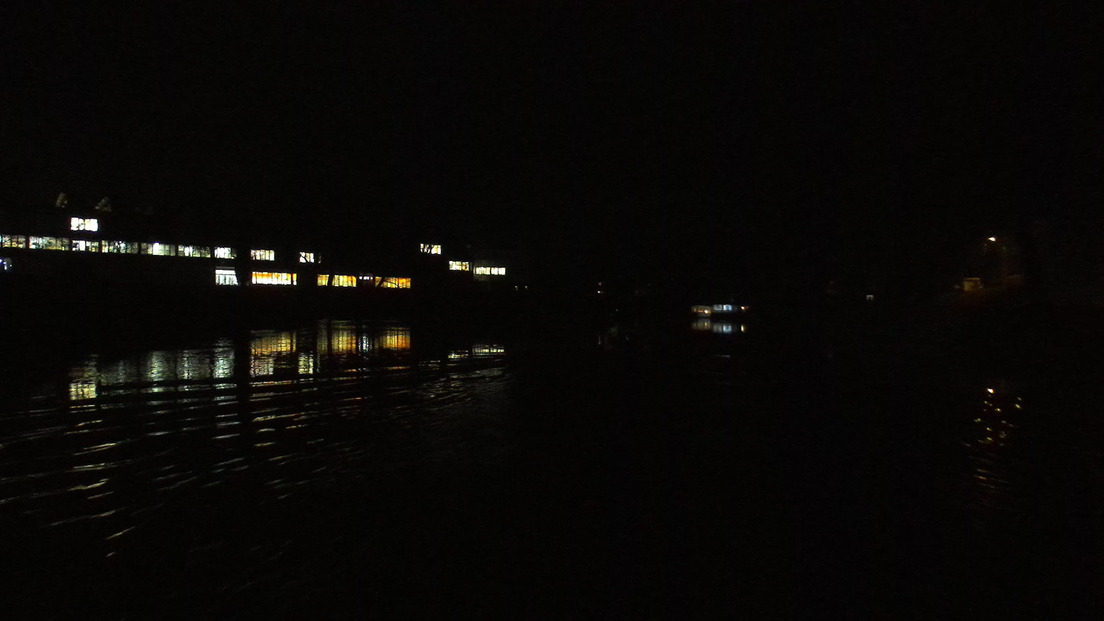}} &
    \frame{\includegraphics[width=\twidth\textwidth]{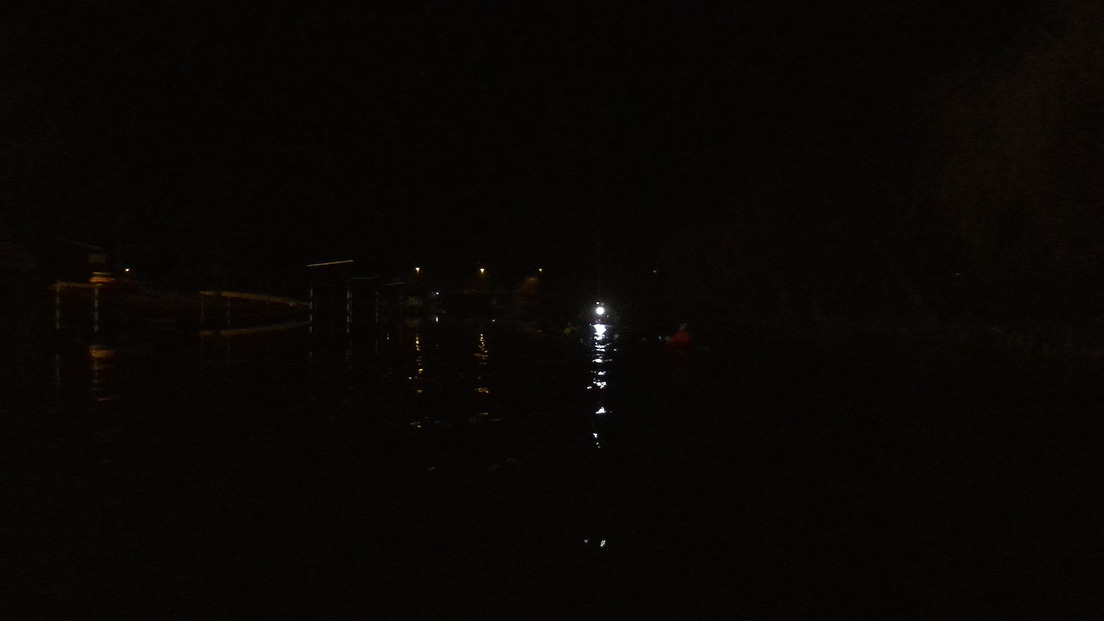}} &
    \frame{\includegraphics[width=\twidth\textwidth]{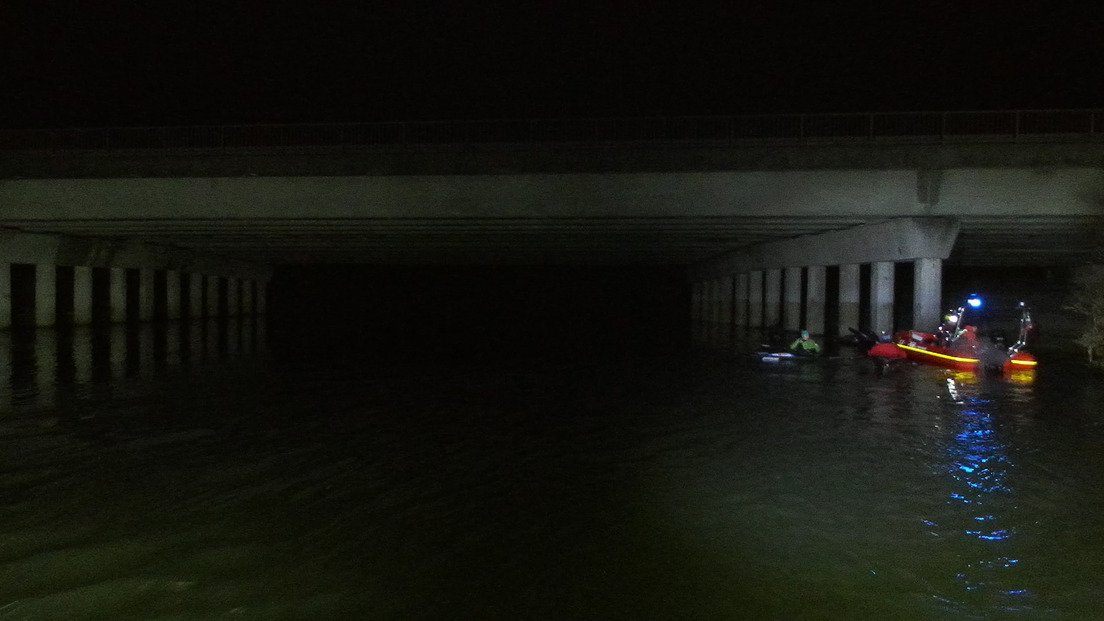}} &
    \frame{\includegraphics[width=\twidth\textwidth]{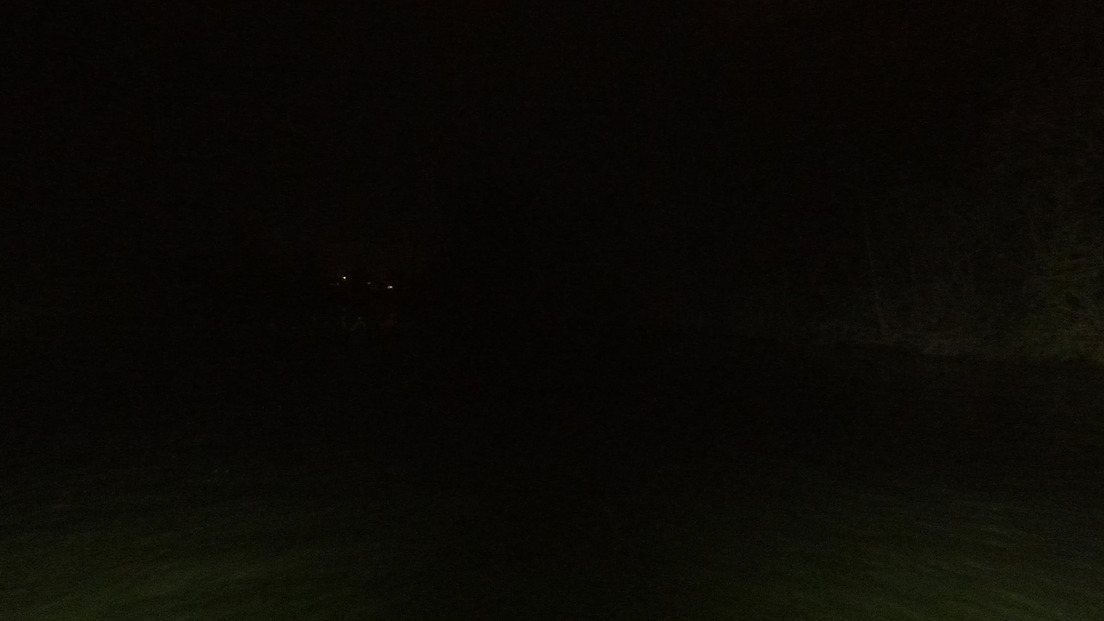}}\\

    \multicolumn{4}{c}{Images captured on the Ljubljanica river at night in non-urban areas (\textit{lj4})} \\[0.5cm]

  \end{tabular}
  \egroup
  
  \caption{Examples of images from our dataset, captured at different locations and under different circumstances. Note the high contrast, sun glare, cluttered environments and low-light conditions.}
  \label{fig:dataset_examples}
  
\end{figure*}

Since datasets used for supervised learning are only as good as the corresponding annotations, special care was taken to ensure high-quality annotations. Human annotators are generally able to annotate image data captured under bright and clear circumstances with little problem. Maritime data can, however, be quite challenging in some edge cases. The edge of water surface can be difficult to determine precisely if the water surface is calm or the border is in the shade. Due to perspective projection, the boundaries between distant objects can be difficult to define due to their small size, and even spotting a small distant obstacle can be challenging due to its visual similarity to waves (c.f. Figure~\ref{fig:dataset_examples}). If looking at the sun, the ensuing glare and increased contrast can cover up even large objects almost entirely, thus the annotator must rely on additional temporal information or data from different sensors to produce valid annotations.

The dataset was annotated with the task of semantic segmentation in mind, using the following classes: \textit{static obstacle}, \textit{dynamic obstacle}, \textit{water} and \textit{sky}. \textit{Static obstacle} denotes the shore, buildings, overhead bridges and overpasses as well as piers and other immovable waterborne objects. The \textit{dynamic obstacle} class was used for movable objects such as boats and paddleboards as well as animals, buoys, swimmers and other objects completely surrounded by water that are visually distinct from it. An auxiliary class \textit{recording boat} was also used during annotation to label static parts of the recording boat visible in some sequences. During supervised training, pixels with this class are ignored. The dataset was annotated per-pixel, using the RGB images captured by the ZED stereo camera. This set of classes was deliberately chosen to be coarse, because clearly determining the class and instance segmentation of faraway objects on the water surface can be very difficult for human observers. Deliberately avoiding closed vocabulary annotations also simplifies the annotation process, as well as possibly lead to better model generalization.

The annotation of our dataset was performed by experienced annotators and supervised by computer vision experts that specialize in the interpretation of maritime environments. The annotators were given the color images, with an optional overlay of projected LIDAR points for reference if the obstacles were not discernible (e.g. top left image in Figure~\ref{fig:dataset_examples}). The images were annotated sequentially, with extra care taken that objects were annotated consistently across the entire time they were in the field of view of the camera. Several iterations of annotation followed by expert critique were performed to consolidate the annotations and ensure the highest possible quality.

\subsubsection{Annotating nighttime data}

The sequence recorded at night (\textit{lj4}) required a different process of annotation, since the color images were mostly too dark to interpret precisely. To alleviate this problem, we chose to use thermal camera images for annotating this sequence. Due to the principle of its operation, the nighttime thermal images are nearly indistinguishable from daytime thermal images. However, due to lower resolution and lack of texture in thermal images, even experienced annotators find thermal images difficult to label. Thus, the annotators were first shown RGB-thermal image pairs from daytime recordings and tasked to annotate the thermal image instead. The LIDAR points were also projected onto both images to help with interpreting the scene. In this task, the color images were used only as support for interpreting how different objects and structures appear in thermal images. After a training period, the annotators were tasked to annotate thermal images taken at night, without the additional color image information.

\subsection{Structure}
\label{sec:structure}

One of the main requirements with multi-sensor systems is the extrinsic calibration. Without it, aligning the sensors and subsequent sensor fusion are nearly impossible. We calibrated our sensor system using the approach presented in~\cite{muhovivc2023joint}, but will briefly outline the process here as well.

The extrinsic calibration procedure was performed on each camera-LIDAR pair independently, with the LIDAR being the central sensor to which other cameras were related. To achieve this, we used two versions of the standard asymmetric grid calibration target, one for visible light spectrum and the other for IR spectrum. The IR target was designed specifically for calibrating the NIR cameras, but also served as a target for the thermal camera. By detecting the target edges in both camera images and LIDAR, a gradient descent optimization process was initialized and the relative position of both sensors was determined. Since the positions of the sensors are fixed, establishing the camera-LIDAR relative position also transitively allowed us to establish relative positions between different camera sensors. As our dataset is focused on semantic segmentation, the main sensor for annotation was the left camera of the ZED stereo system, which also has the largest field of view of all included cameras.

For producing pixel-aligned multimodal data frames, data from other sensors had to be transferred to the ZED image plane. This was straightforward for LIDAR, but required extra work for other sensor images. Due to the parallax effect, the image pixels cannot be directly related unless the scene is at infinite distance, thus additional depth data is required. Since ZED is a stereo camera system, we were able to use the dense stereo depth to perform the image-to-image projection and thus bring all the different camera modalities to a common image plane. Since the quality of the stereo depth is reliant on the ambient lighting, the stereo data was not useful for nighttime data. Thus, LIDAR data was used for calculating the approximate dense depth values. The 3D points were projected onto the ZED image plane as control points, then radial basis function interpolation was performed to produce dense depth approximations for the entire image plane. Examples of thermal images remapped in this manner are shown in Figure~\ref{fig:title} and Figure~\ref{fig:results}.

\noindent The dataset is synchronized to the ZED camera timestamps, i.e. for each timestamp we find the data that was acquired closest to the current timestamp. If the difference is smaller than the sensor sampling period, we consider the data valid. This can incur some small displacements because of the asynchronous nature of the sensor system. However, due to the relatively low dynamics of the maritime domain, this causes no noticeable issues in our data. The following data from all sensors at discrete timestamps is available:
\begin{itemize}
\itemsep0.3em
    \item ZED RGB, $\SI{2208}{\px} \times \SI{1242}{\px}$, 24-bit, encoded as JPEG images
    \item ZED depth, $\SI{2208}{\px} \times \SI{1242}{\px}$, 16-bit
    \item IR1, $\SI{896}{\px} \times \SI{1080}{\px}$, 8-bit, encoded as JPEG images
    \item IR2, $\SI{896}{\px} \times \SI{1080}{\px}$, 8-bit, encoded as JPEG images
    \item Polarization images, $\SI{1216}{\px} \times \SI{1024}{\px}$, 32-bit (four 8-bit polarization channels), encoded as RGBA PNG images
    \item Thermal images, $\SI{384}{\px} \times \SI{288}{\px}$, 16-bit, encoded as PNG images
    \item Radar point cloud, $N\times 5$ ($x$, $y$, $z$, \textit{speed}, \textit{RCS})
    \item LIDAR point cloud, $N\times 4$ ($x$, $y$, $z$, \textit{reflectivity})
    \item GPS/IMU data, (\textit{latitude}, \textit{longitude}, \textit{altitude}, \textit{roll}, \textit{pitch}, \textit{yaw})
\end{itemize}

All JPEG encoding has been done at JPEG quality level of 95. The number of points in LIDAR point clouds ranges up to 69k (theoretical maximum for dual return on VLP-16), but can be significantly lower depending on the observed scene. The maximum number of radar points is 110.
Along with GPS data we provide the estimated accuracy, with the best being around 30cm, due to using RTK correction.

For cameras, intrinsic calibration parameters are provided to enable image rectification. Relative position to the LIDAR sensor is also included for camera sensors in the form of a rotation matrix $R$ and a translation vector $t$. Relative (measured) positions of GPS and radar sensors to the ZED camera are also provided. Included as part of the dataset is the code that implements the pixel mapping method presented in~\cite{muhovivc2023joint}. The methods for constructing dense depth images from LIDAR data are also supplied.

\subsection{Dataset splits}

Our dataset is split in multiple different ways to facilitate training and evaluating the model performance in several directions. All of them include training, validation and testing subsets. Our main goal was to provide multimodal maritime data acquired under different circumstances, and to set up the subsets in a way that decouples attributes such as lighting conditions and location.

\subsubsection{Day-night split}
\label{sec:day-night}

The main dataset split focuses on model performance at night, while being trained only on daytime data. The day-night split of the dataset consists of subsets for training, validation, and testing. The daytime data is split into training and validation subsets, which contain per-pixel annotated RGB images taken from all the daytime sequences. When selecting data for annotation, special focus was on including variations in lighting and the obstacles visible in the images. The data was recorded during daytime but under different weather conditions (sunny, cloudy, sunset). The daytime data was split into training and validation sets with the ratio 90:10, where the sampling was uniform.

\begin{figure*}[ht]
    \centering
    \def\twidth{0.16}

    \setlength{\tabcolsep}{0.1em}
    \begin{tabular}{cccccc}
    
    \includegraphics[width=\twidth\textwidth]{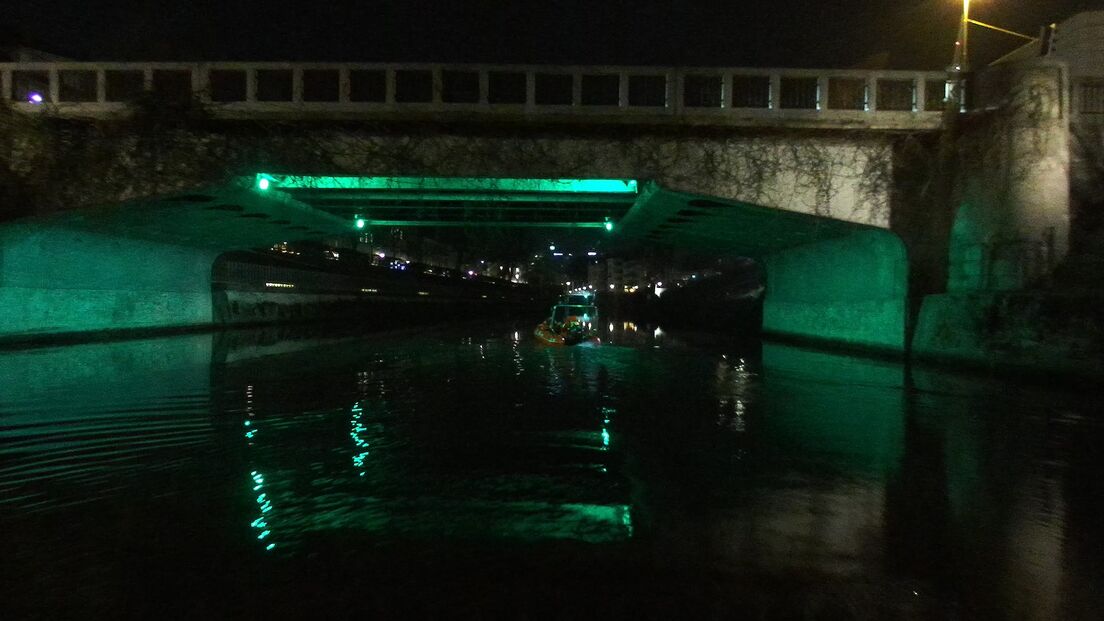} &
    \includegraphics[width=\twidth\textwidth]{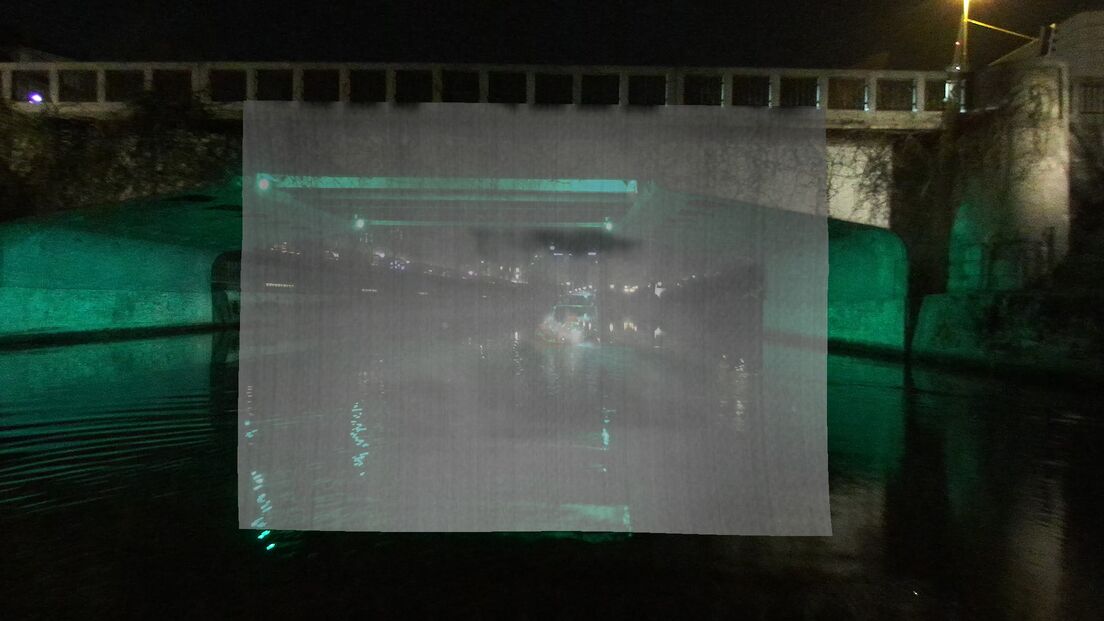} &
    \includegraphics[width=\twidth\textwidth]{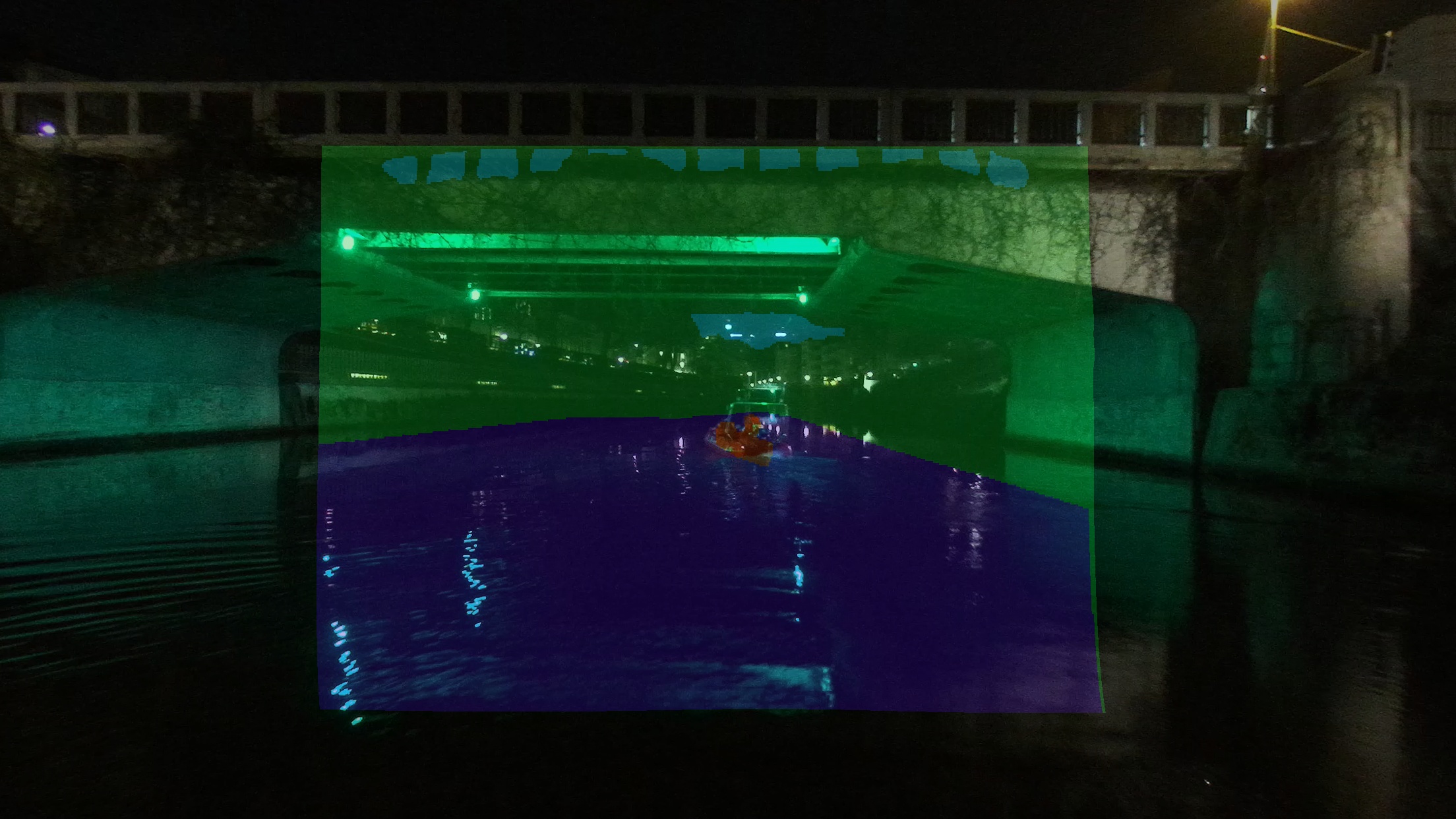} &
    \includegraphics[width=\twidth\textwidth]{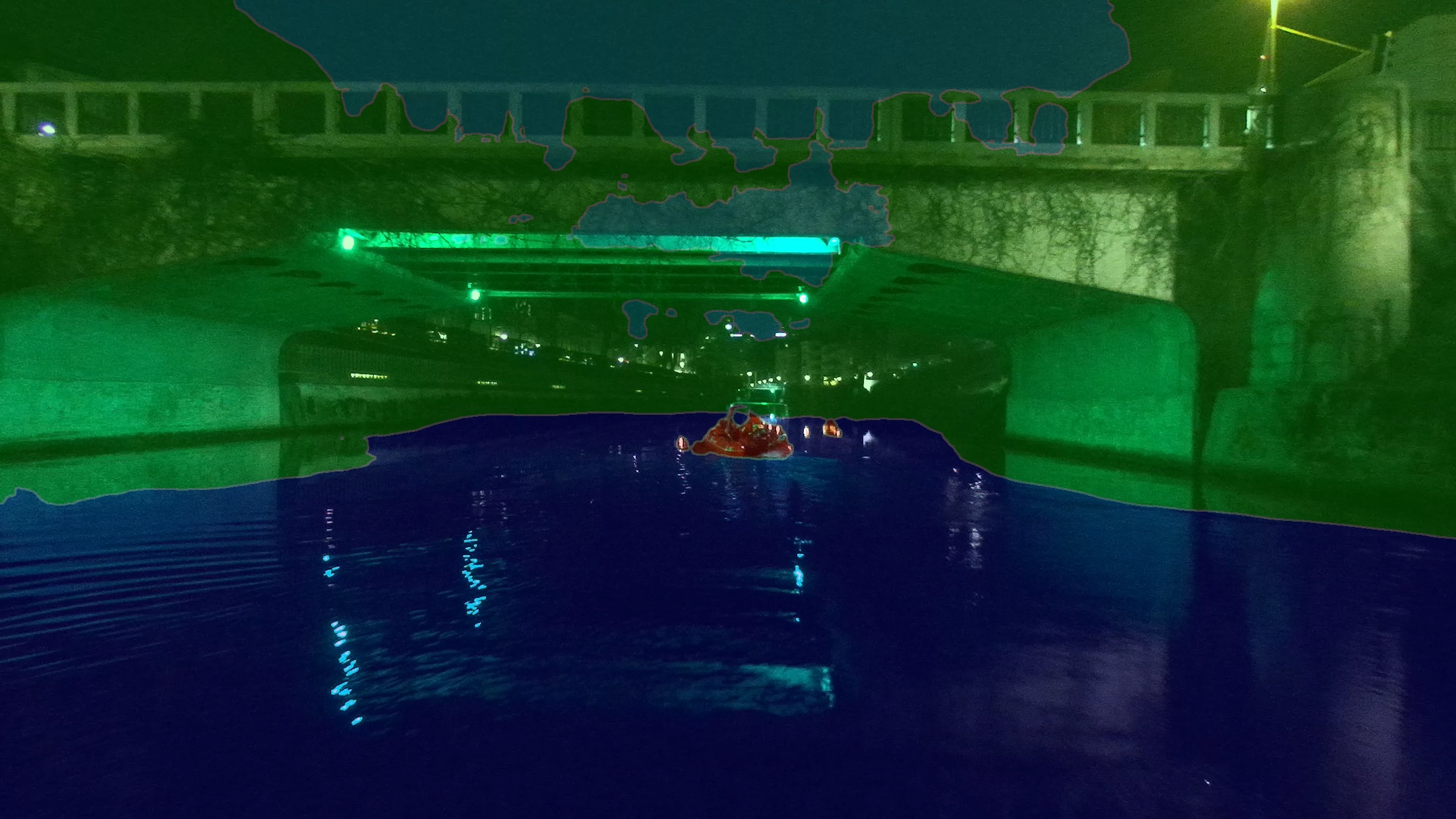}&
    \includegraphics[width=\twidth\textwidth]{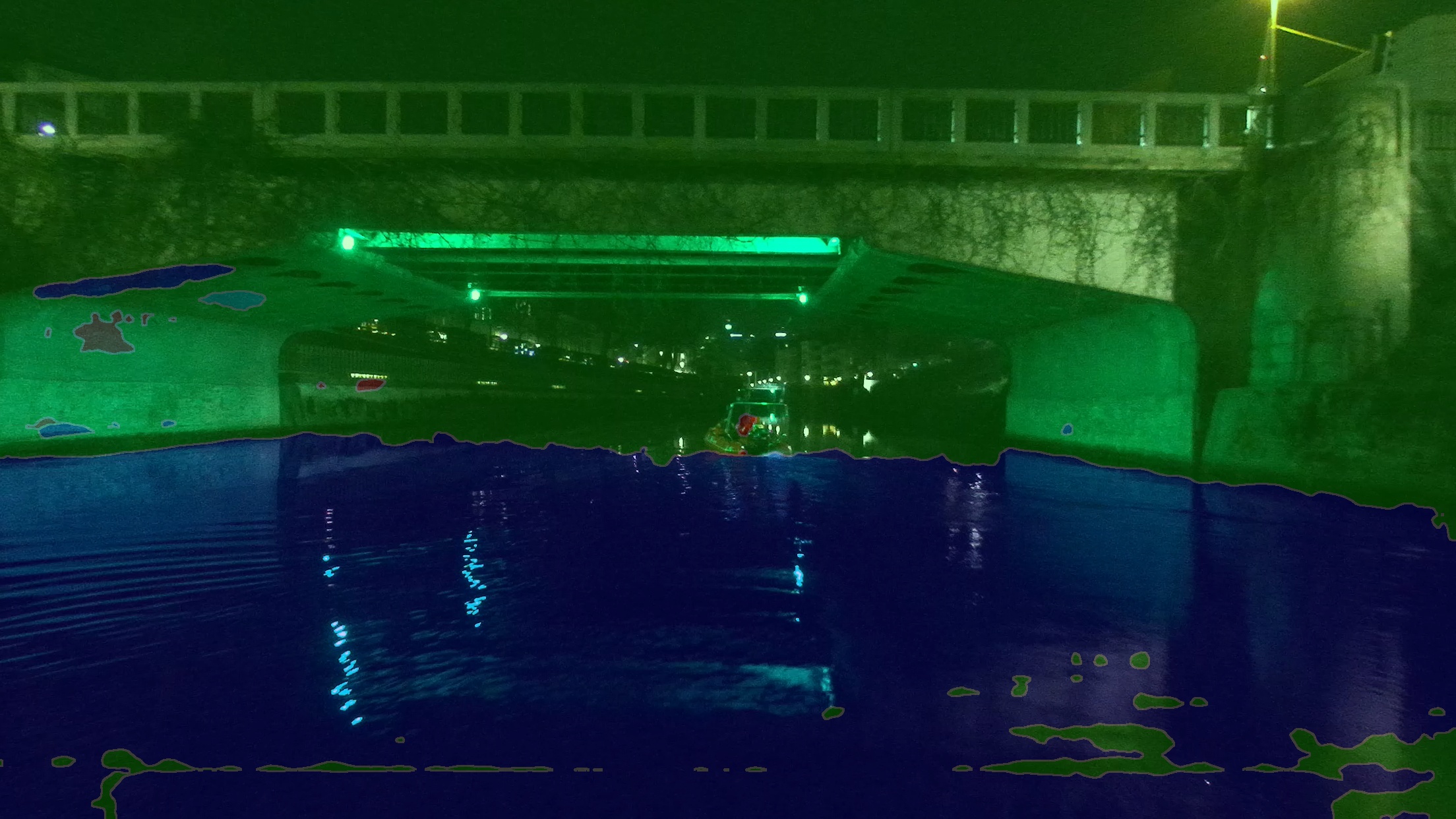}&
    \includegraphics[width=\twidth\textwidth]{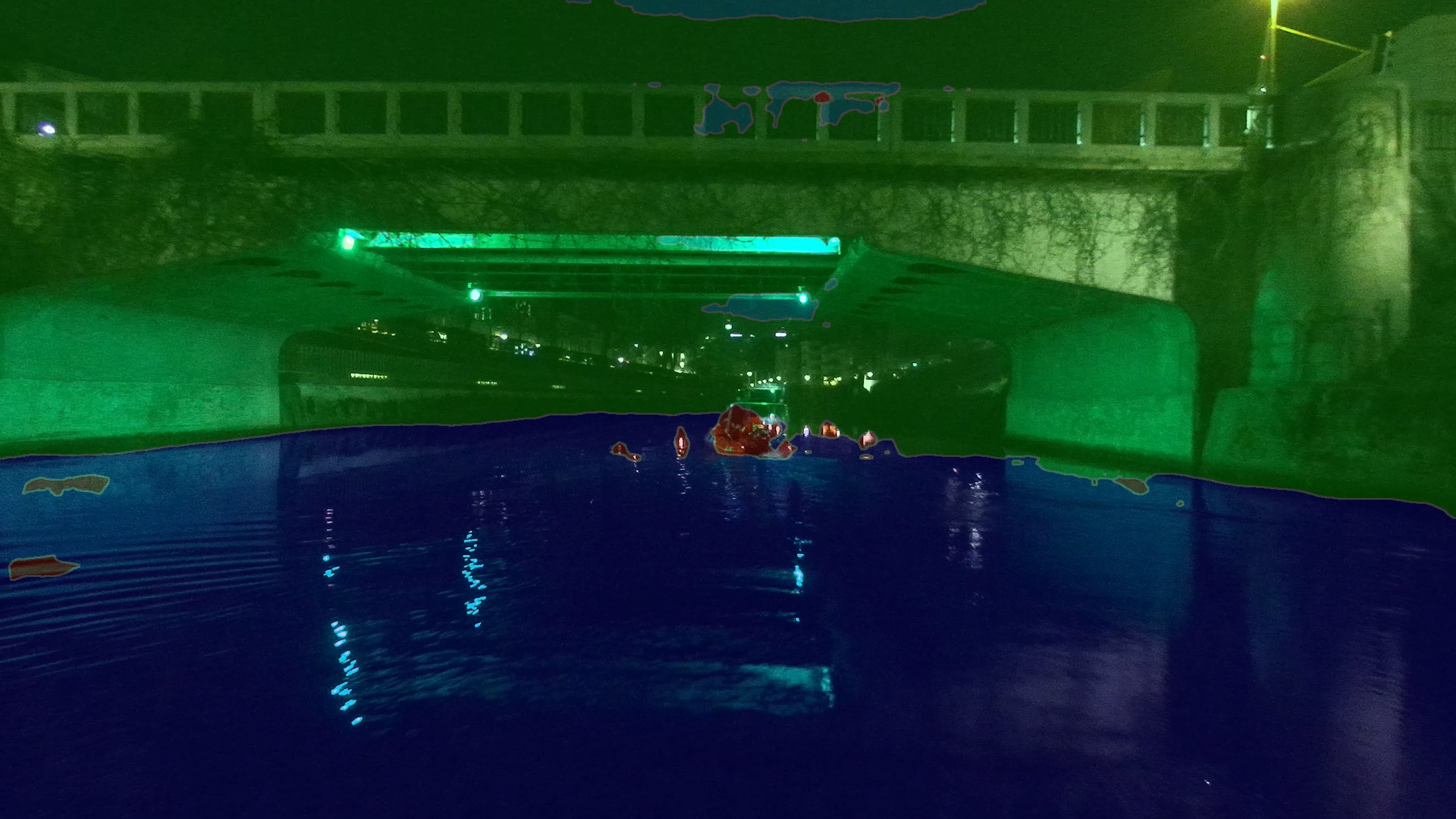}
    \\

    \includegraphics[width=\twidth\textwidth]{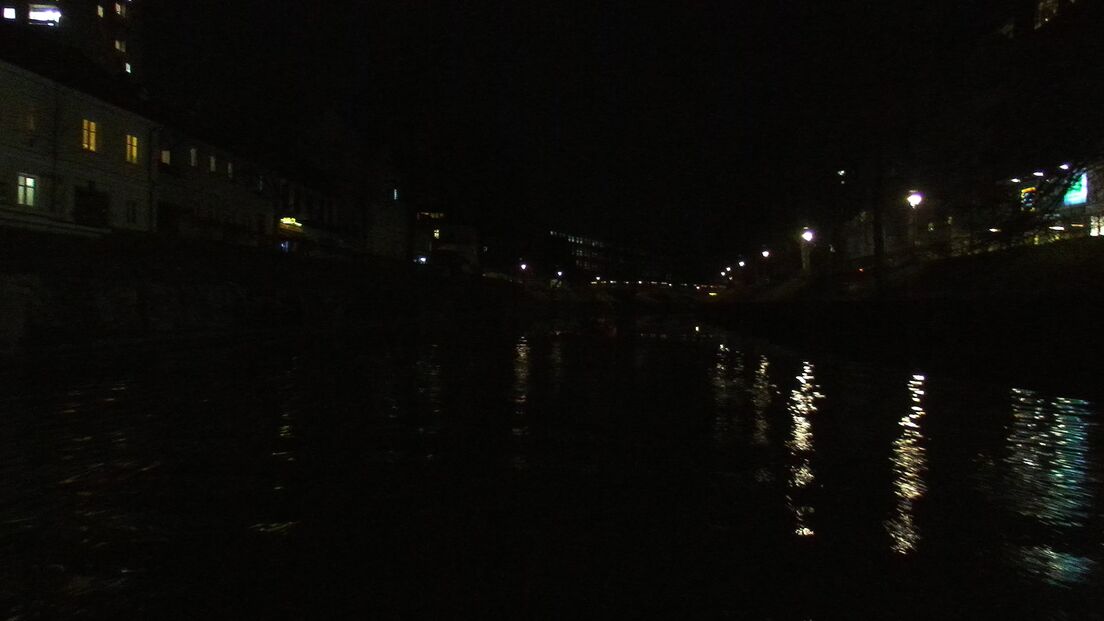} &
    \includegraphics[width=\twidth\textwidth]{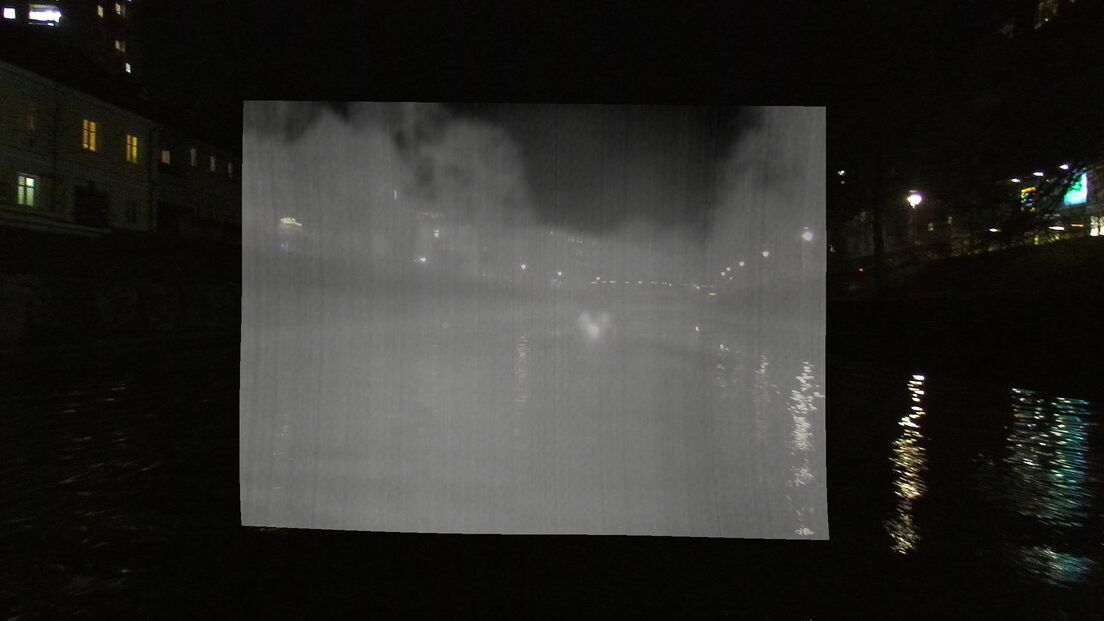} &
    \includegraphics[width=\twidth\textwidth]{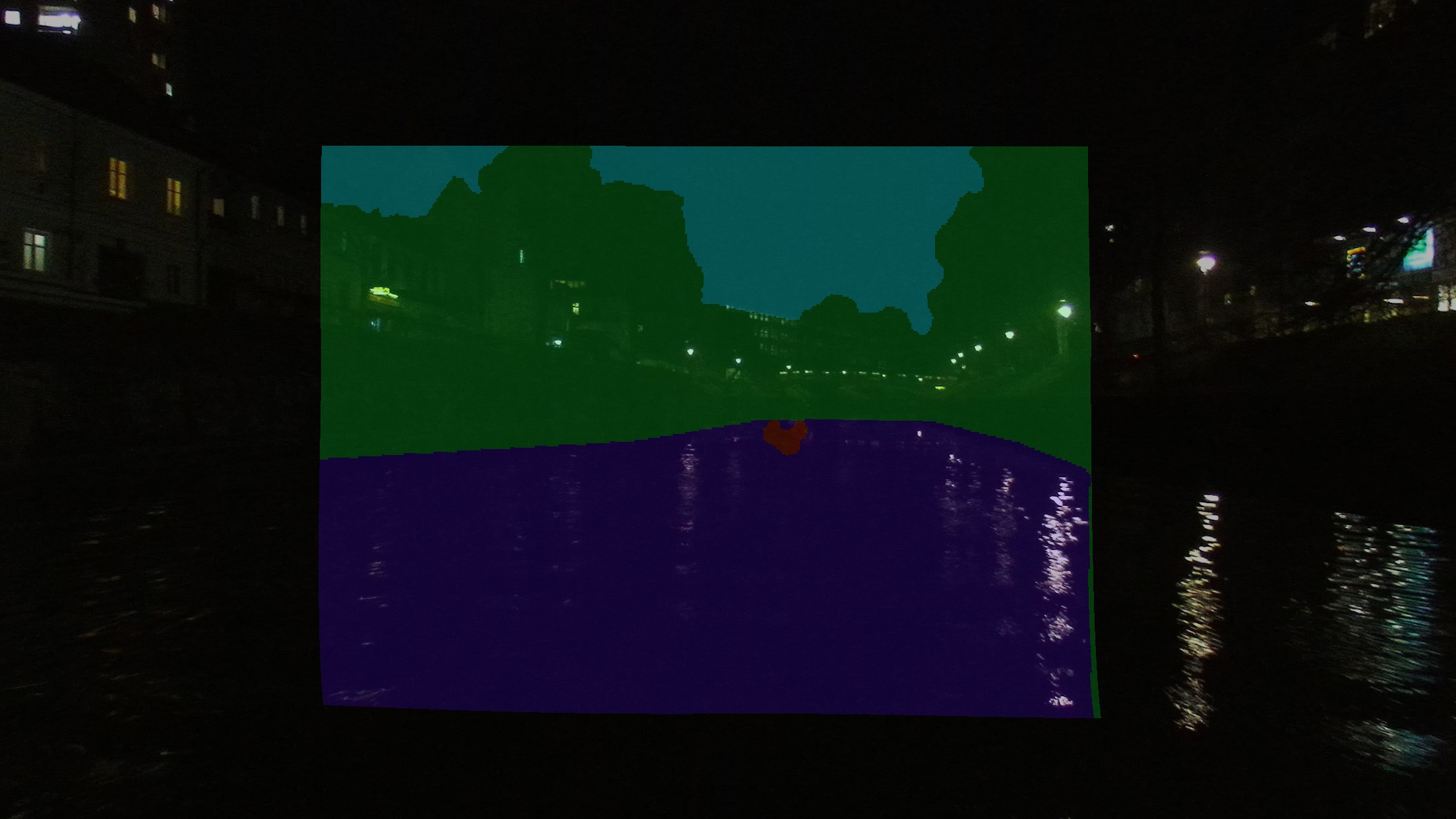} &
    \includegraphics[width=\twidth\textwidth]{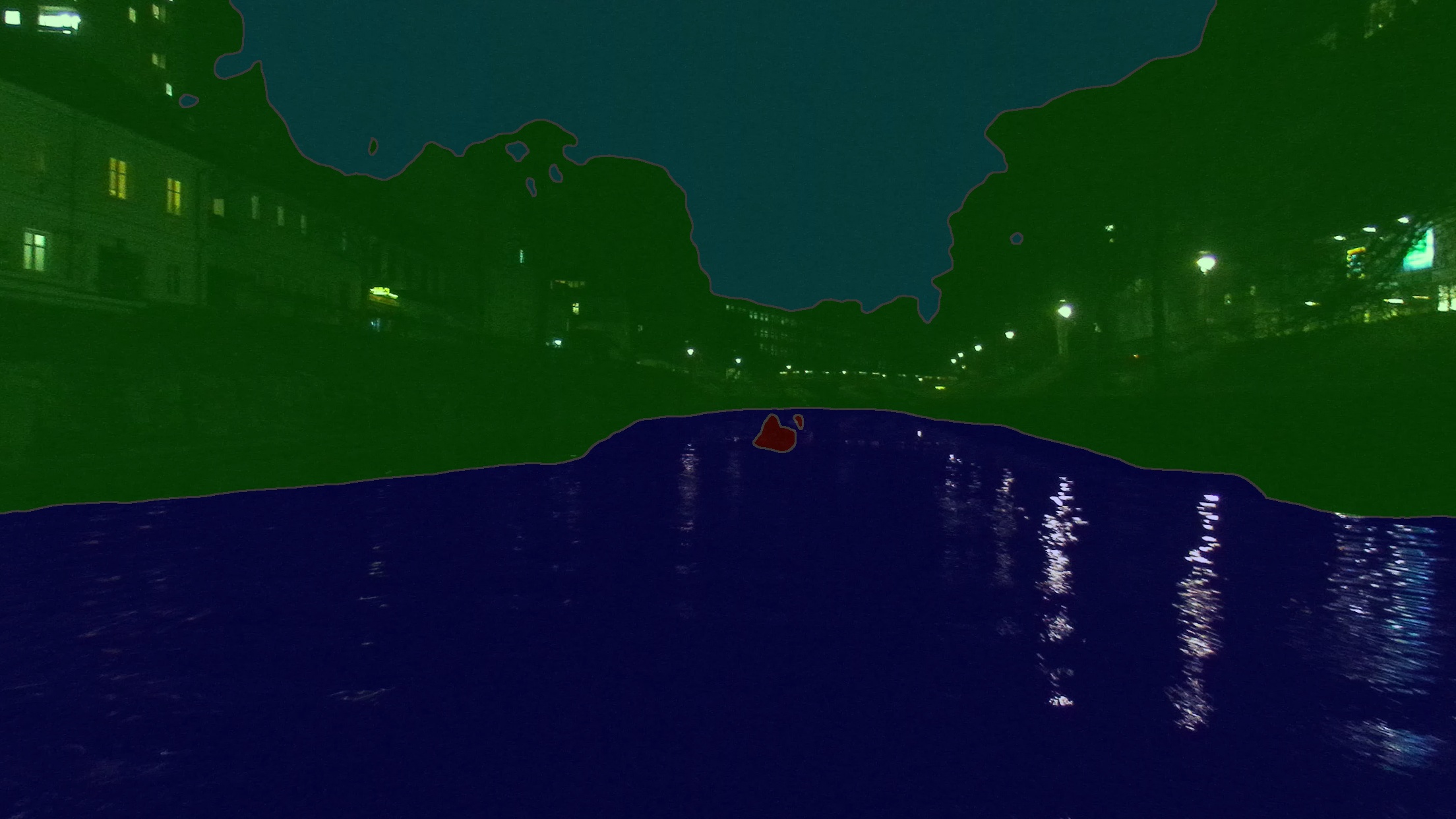}&
    \includegraphics[width=\twidth\textwidth]{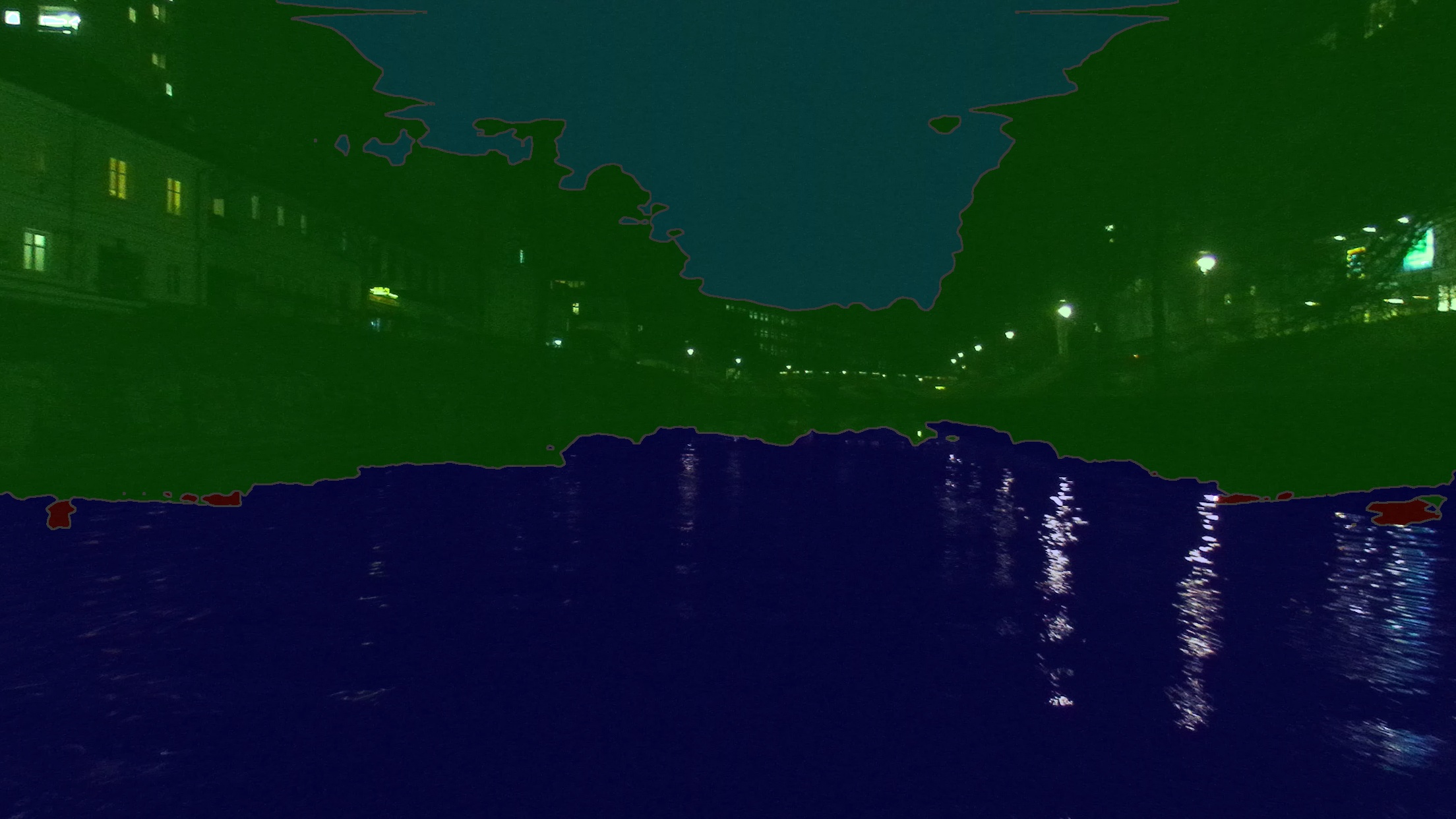}&
    \includegraphics[width=\twidth\textwidth]{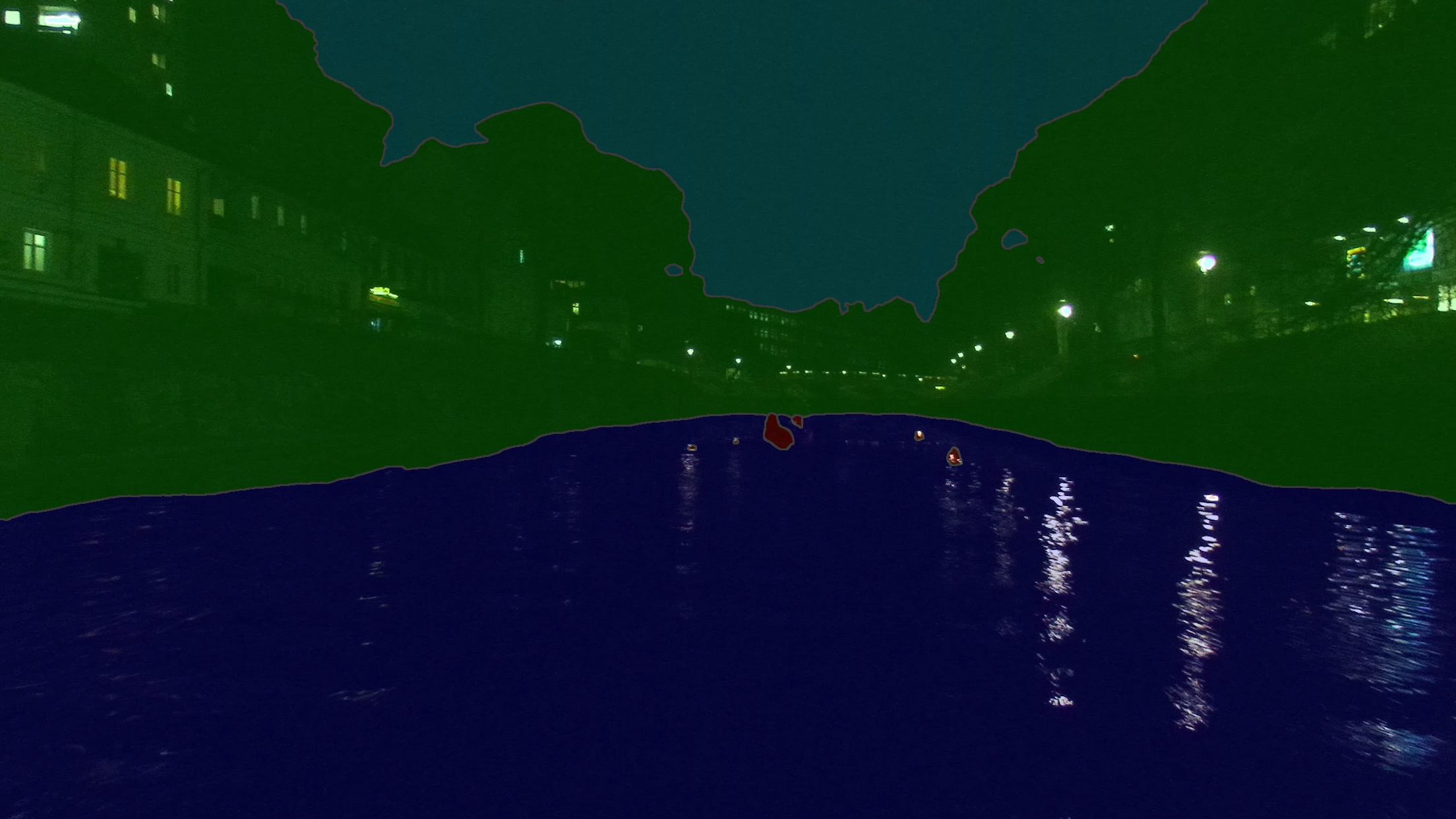}\\

    \includegraphics[width=\twidth\textwidth]{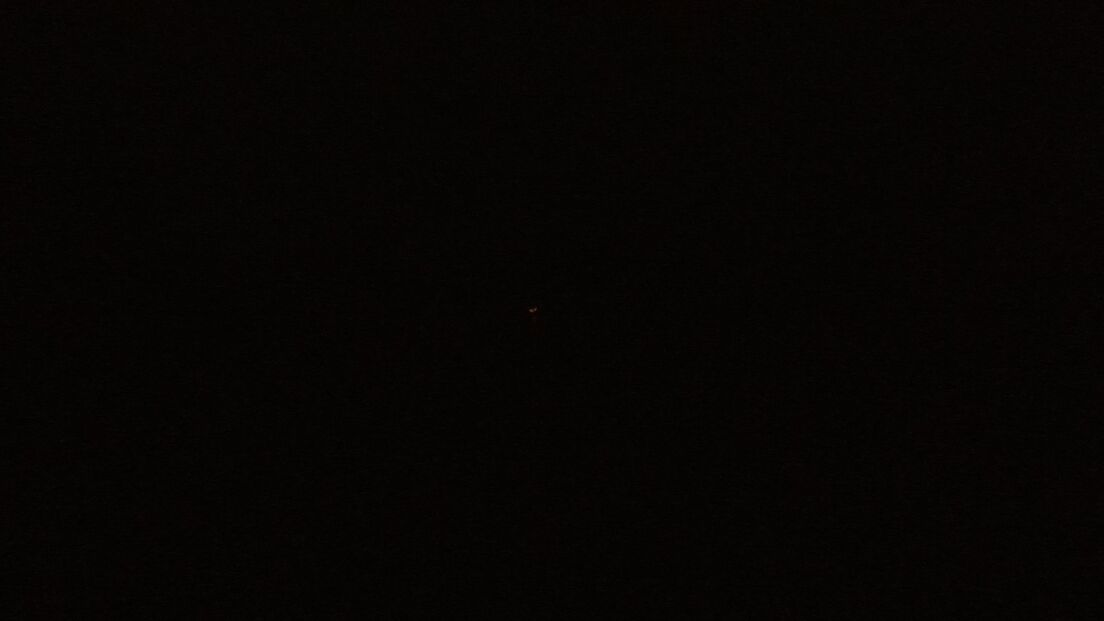} &
    \includegraphics[width=\twidth\textwidth]{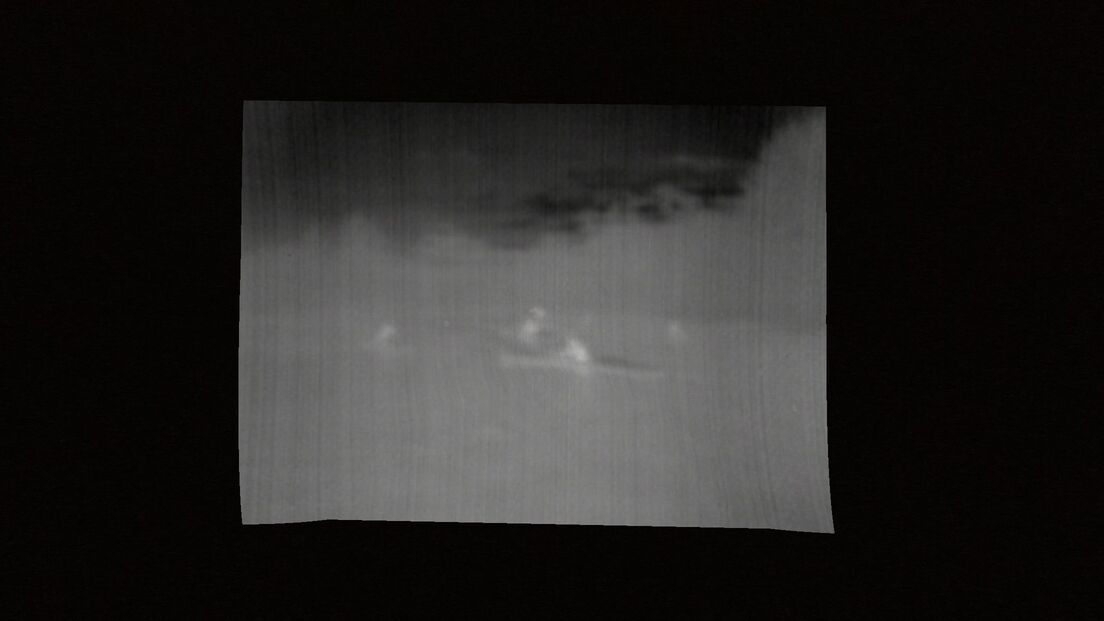} &
    \includegraphics[width=\twidth\textwidth]{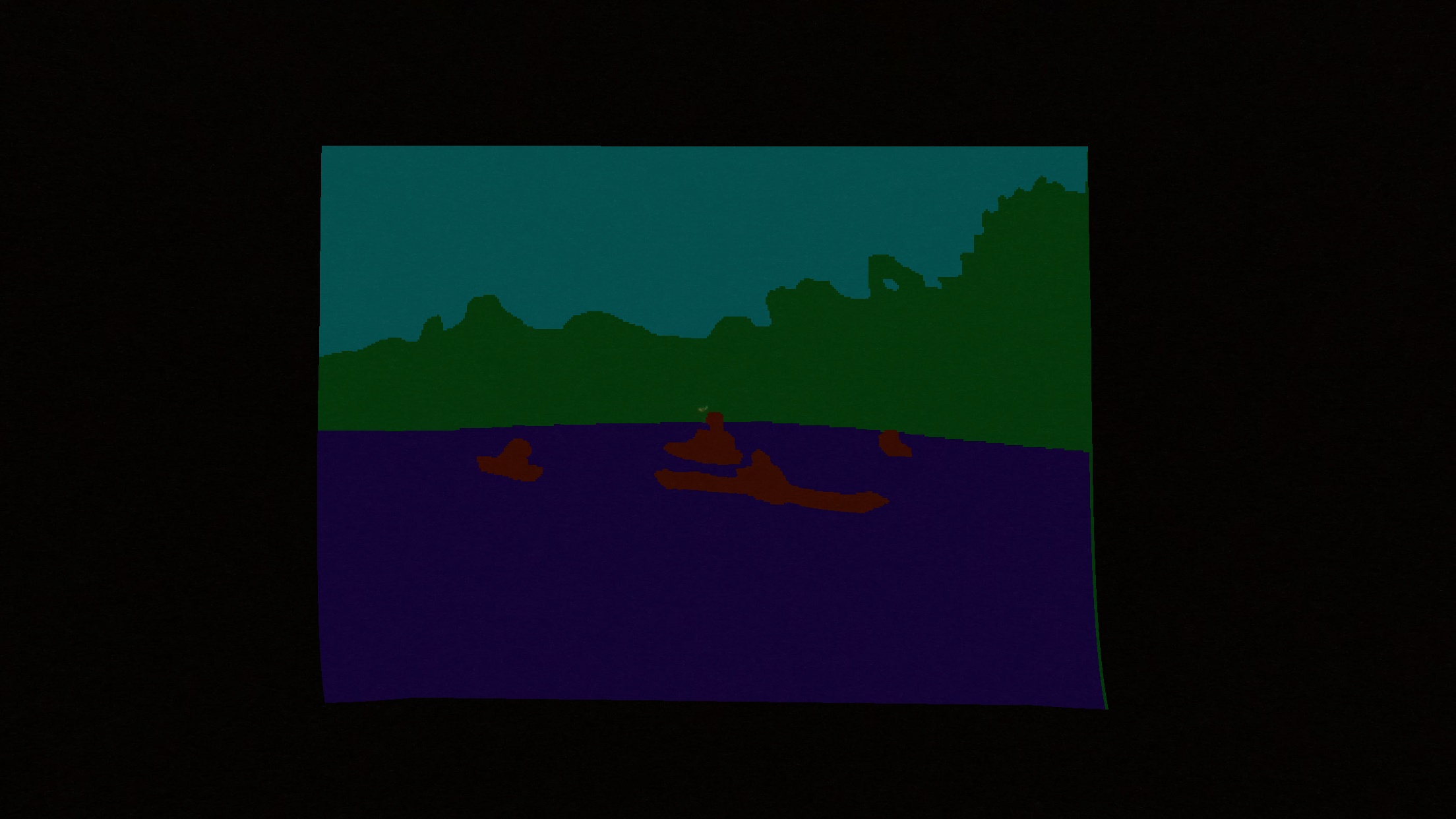} &
    \includegraphics[width=\twidth\textwidth]{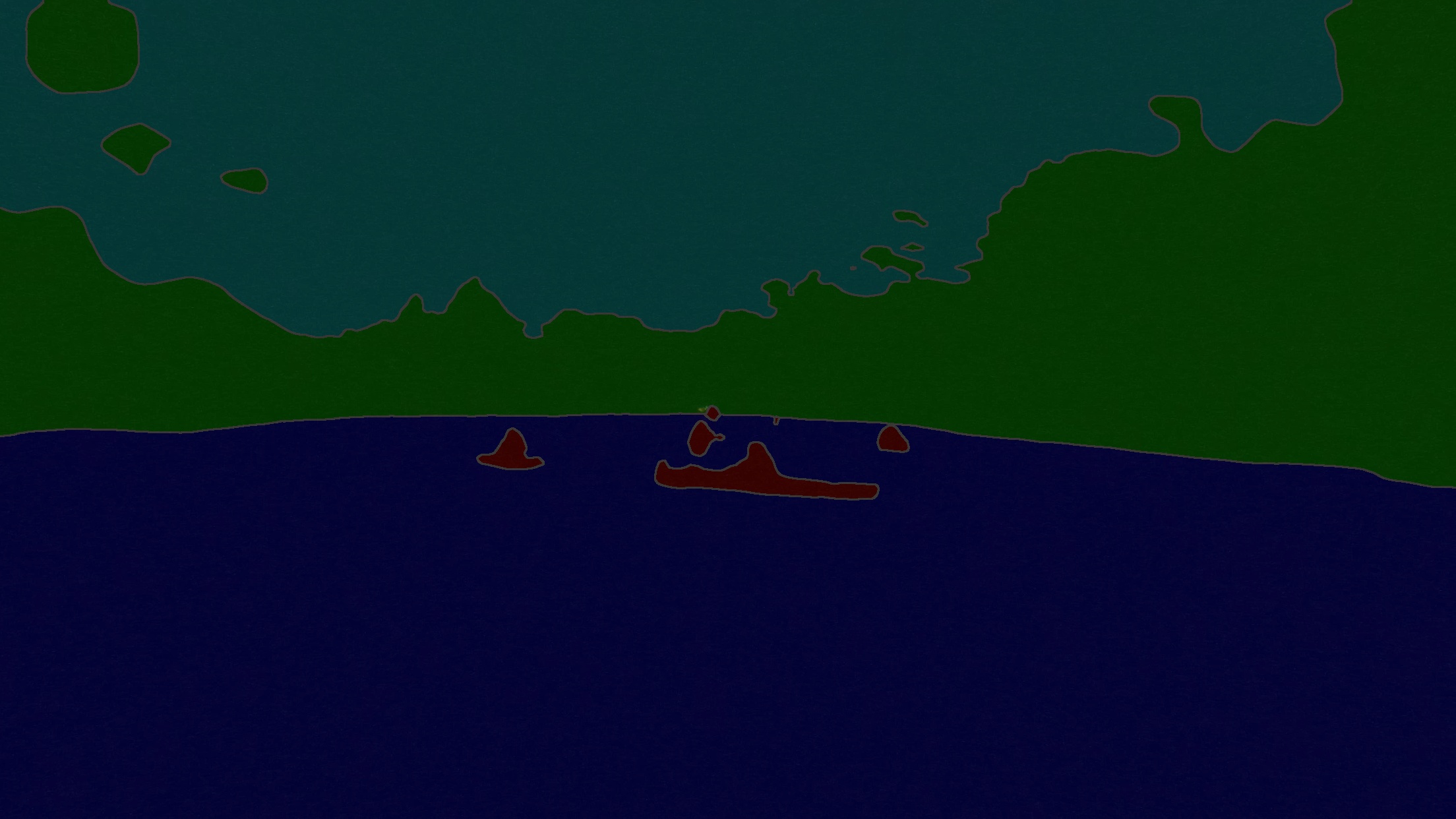}&
    \includegraphics[width=\twidth\textwidth]{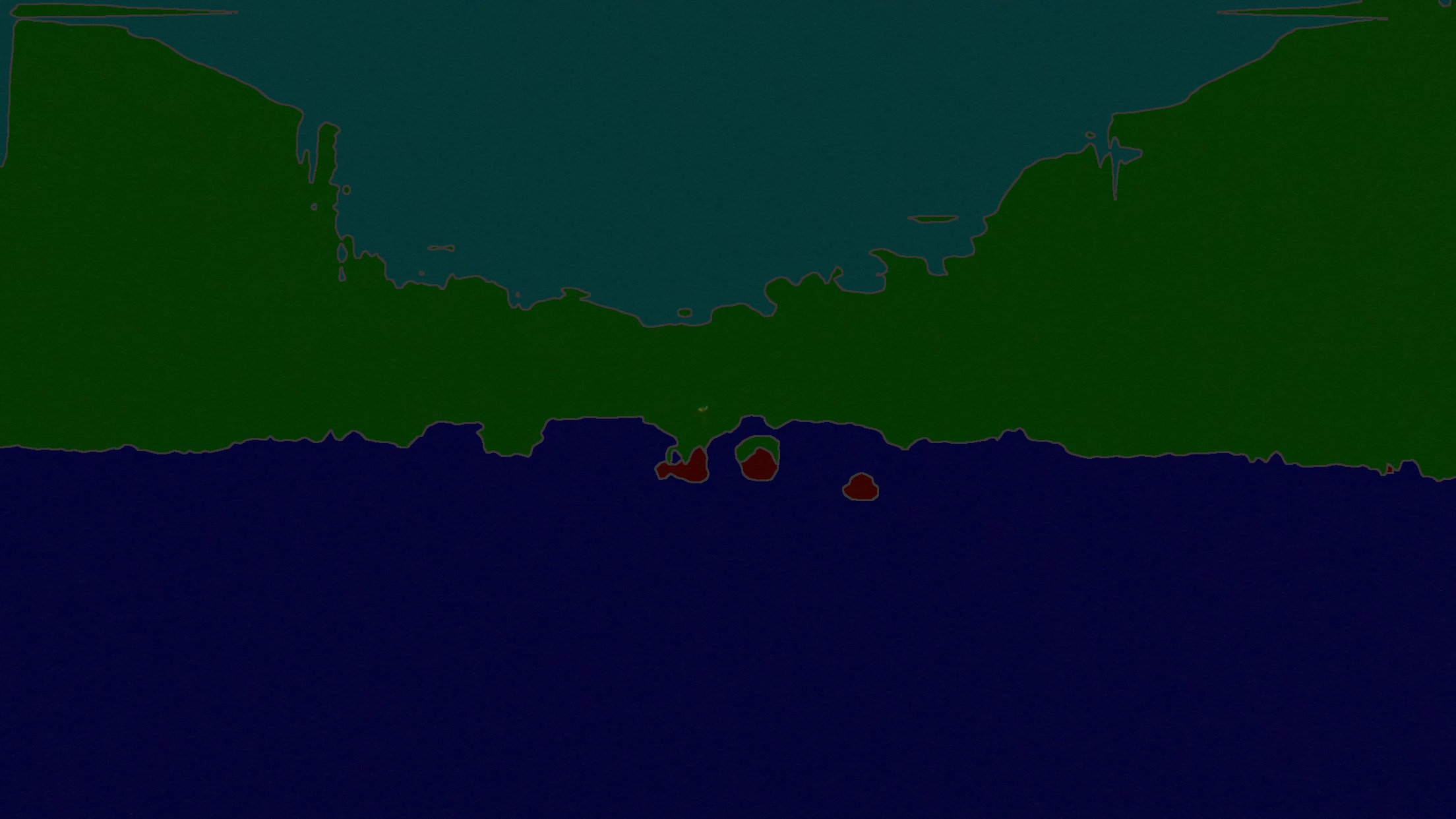}&
    \includegraphics[width=\twidth\textwidth]{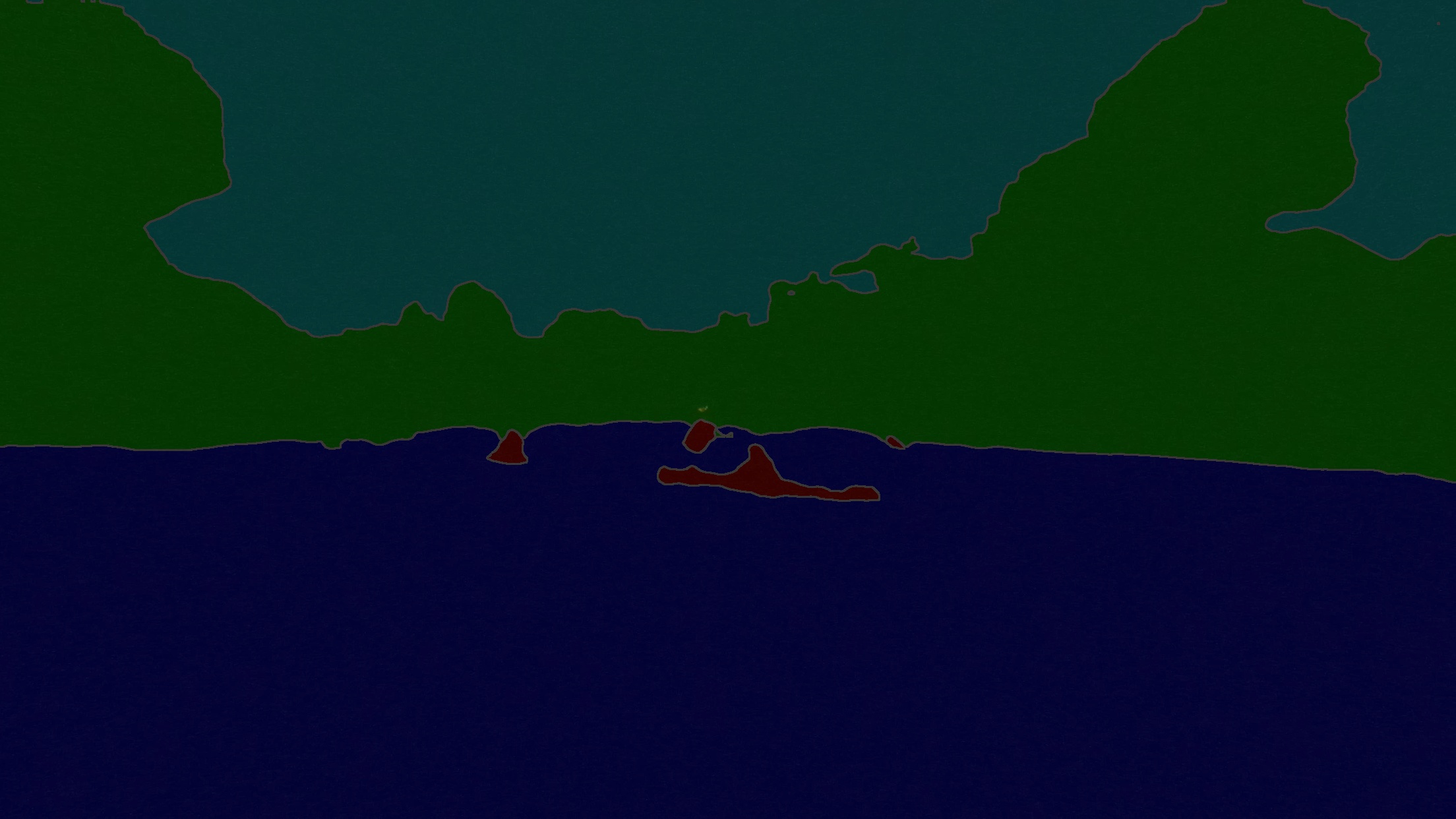}\\

    \scriptsize{RGB} & \scriptsize{Thermal} & \scriptsize{GT} & \scriptsize{CMNeXt-DH} & \scriptsize{MMSFormer-D} & \scriptsize{StitchFusion-D} \\

  \end{tabular}
  
  \caption{Example images from the nighttime test set. The first column shows raw RGB images, the second column shows thermal images overlaid on RGB images, and the third column shows ground truth annotations. The last three columns show the predictions of best-performing nighttime models: CMNeXt-DH, MMSFormer-D, and StitchFusion-D. Semantic labels for \textit{sky}, \textit{static obstacle}, \textit{water} and \textit{dynamic obstacle} classes are shown in cyan, green, blue and red, respectively.}
  \label{fig:results}
  
\end{figure*}

The test subset is a difficult testing scenario and consists of exclusively nighttime data from sequence \textit{lj4}. Due to poor lighting conditions, reflections and sensor noise, this data is very difficult to interpret only from visual light-based images. Example images can be seen in Figure~\ref{fig:results}. However, due to the presence of modalities that are largely unaffected by such conditions (LIDAR, radar, thermal camera), interpretation of nighttime data is still possible, though much more difficult than for data captured at daytime. The test set serves as a benchmark of the capability of a multimodal method to exploit auxiliary modalities present in the dataset. Thus, if one source of data is compromised, the models should be able to still perform reliably based on the remaining data, albeit at a reduced accuracy.

\subsubsection{Alternative evaluation splits}

We provide three additional data splits that can be used for evaluating the generalization capabilities of methods and the effect of using auxiliary data on difficult daytime examples. The ability of methods to generalize to different locations can be tested by using the \textit{geography} split of our dataset, where only data captured on the Ljubljanica river is used for training, and data from other locations is only used for testing the final model. The validation subset is uniformly sampled from training sequences \textit{lj1}, \textit{lj2}, and \textit{lj3}. The \textit{saltwater} split sequesters the images captured on the seaside to measure the model's generalization to different types of water surfaces. This split uses data captured on the river and on the lake for training, whereas data from sequence \textit{adr1} is only used for testing. In order to gauge the usability of auxiliary modalities, the \textit{difficult} split removes difficult parts (i.e. those that cannot be reliably solved just using RGB images) of the daytime data from the training set and uses them as part of the test set. For example, the test set of this split contains data with strong sun glare and reflections that can cause RGB images to be difficult to interpret. Using auxiliary modalities (polarization or thermal camera) can alleviate such issues, and this split can serve as a benchmark for such cases. The validation set in this case is uniformly sampled from non-difficult parts of daytime sequences, whereas the test set only contains difficult parts of captured daytime sequences.

\section{Method}
\label{sec:method}


The field of multimodal methods that use additional image modalities to improve performance for image interpretation tasks has been developing significantly lately. Many methods were proposed that produce predictions based on RGB images complemented by an extra modality (such as depth), but fairly few of them support more than one additional modality. This stems from the increased complexity when calculating inter-modality information exchange if many modalities are used. Authors propose several approaches for limiting the increase in calculation complexity, such as merging auxiliary modalities early or using shared weights for different modalities.

The authors have shown that multimodal methods are able to use additional information to produce better predictions when compared to only using color images. This is shown clearly across many modalities and datasets. However, it turns out that either by architecture design or by the nature of used datasets, the methods are heavily biased towards using information from color images. This is reasonable, since the color images are in most cases the most informative of the modalities, carrying dense information about textures, edges, and lighting conditions in the observed scene. Other modalities can include different information, but they are mostly more specialized towards temperature, movement, or distance (as in the case of thermal, event or depth cameras/LIDAR). The consequence of this bias is that such multimodal methods require the RGB channel to be of high quality at inference time, and can thus produce good predictions only if this is true. 

The overreliance on color images can cause issues in multimodal systems used in real environments, where any of the used sensors can potentially either fail to deliver data or the data captured is of poor quality. Since several of the sensors included in our system are not affected by the quality of lighting in the observed scene, we expect that multimodal methods using these modalities would be able to produce quality predictions from them even if color image data is poor. As will be shown in the Section~\ref{sec:experiment}, none of the methods are able to capture useful information from auxiliary modalities if the `main` modality that is degraded or missing. We propose several changes in the architecture and training process that allow the methods to perform better under difficult circumstances.


We used the CMNeXt~\cite{zhang2023delivering} architecture as the basis for our architecture and training adjustments, but the changes are general enough that they can be applied to most other architectures as well. CMNeXt is an approach to multimodal semantic segmentation that uses a mix-transformer backbone for feature extraction and is able to include an arbitrary number of auxiliary modalities to improve the performance of the model. The auxiliary modalities are processed in each layer of the encoder and informative features are selected and fused with the features extracted from RGB data. After encoding, the fused feature vectors are upscaled and predictions are calculated using a simple decoder head, as per the SegFormer~\cite{xie2021segformer} approach. This approach allowed the authors to achieve several state-of-the-art results, on popular multimodal datasets.

\subsection{Architecture adaptation}

Our work initially focused on determining which modalities could help the model interpret the observed scene. We chose to explore the hardest possible task, that being the case when color image data is poor or almost entirely unusable. We selected three modalities from our dataset: RGB, thermal camera and LIDAR. The auxiliary modalities were chosen due to them being largely unaffected by large changes in lighting conditions. This makes them suitable for providing information about the environment during nighttime or other low-light scenarios. Due to the difficulty of obtaining and annotating nighttime data, we decided to only use daytime data for supervised learning. Note that the design of a model that does not rely on training in difficult circumstances is of key importance. Data acquisition in extreme conditions can be dangerous, and producing annotations in large enough quantities can be very difficult.

Using the same model for both daytime and nighttime scene parsing can be viewed as a domain adaptation problem, where the domain used for training is daytime data and the target domain for which annotations are not available is nighttime data. The design of our method is similar to what the authors of HeatNet~\cite{vertens2020heatnet} proposed. However, in contrast to using pretrained teacher models, we choose to train our model in an end-to-end manner and guide the training process so that the fusion modules can learn to dynamically focus on the most informative modalities needed to correctly parse the observed scene. If the feature extraction is trained well, the model should be able to use the unaffected modalities to perform semantic segmentation even if RGB images contain little information.

\subsubsection{Double forward pass}
Training the CMNeXt architecture only on daytime images along with auxiliary modalities showed that the performance of the data fusion module depends highly on the training data. While auxiliary data is informative enough to be used on its own, the network is not incentivized to use it if RGB data is good enough for quality segmentation. This same observation was the main motivation for the CRM approach~\cite{shin2023complementary}. In order to force the architecture to exploit additional modalities regardless of information contained in RGB images, ideas presented in CRM were used to improve the performance under difficult circumstances. In CRM, the authors showed that removing parts of input images when training for semantic segmentation using both RGB and thermal data improves performance overall, since the network is forced to extract information from both available modalities. This allowed the method to achieve state-of-the-art results on MFNet dataset~\cite{ha2017mfnet}.

Inspired by the self-distillation loss used in CRM~\cite{shin2023complementary} and by the training approach described in the HeatNet paper~\cite{vertens2020heatnet}, we propose to train the network using two forward passes in each iteration. The first pass is performed using all the available data (RGB, thermal, LIDAR) and serves to train the network for daytime use. The second pass is performed using the exact same data, but setting the RGB input to zero, thus also guiding the network towards using the auxiliary modalities. This incentivizes the network to learn and use high-quality features from auxiliary modalities, since rich RGB information is unavailable. Additionally, the fusion module can thus focus on only learning to differentiate between inputs with or without usable RGB data and focus only on informative modalities for the final predictions.

The choice of input for the second forward pass is based on the domain knowledge of our problem. In our case, we set the RGB input to zero, to simulate 
the extreme version of the situations in which RGB images are unreliable (i.e. nighttime or sensor unavailability). The input characteristics could be extended for different conditions under which RGB is known to perform poorly, such as fog or rain. Here we show that even a very rough approximation of difficult conditions can lead to the model learning to interpret auxiliary modalities by themselves, and that explicitly training on difficult data samples is not necessary, since both the data and the annotations can be very hard to obtain for real-life datasets. Additionally, this approach can be extended to cover the absence or poor performance of other sensor modalities as well, since any of the modalities could be synthetically degraded or removed during training to improve the resilience of the final model.

\subsubsection{Modality-specific decoder heads}
We observe that CMNeXt produces multimodal features in each stage that are used for high quality semantic predictions. However, because of feature fusion at each encoder stage, these features are tightly coupled and are only trained in conjunction. This works well when all data is available and of high quality, but performs poorly if data from a sensor is missing completely or is of poor quality. This can be observed by masking each of the input data sources and performing inference, as shown in Table~\ref{tab:ablation}. A large drop in prediction quality can be observed if even one of the modalities is missing. To address such situations, the feature extraction and feature fusion should be less interdependent during training.

In order to achieve this, we added two additional decoder heads to the decoding stage of the model. The features produced by both RGB and auxiliary modalities are gathered from the encoding layers and passed to specialized decoder heads. During training, these heads are guided towards producing good predictions on their own, thereby improving the performance of the network by decoupling the feature extraction and fusion parts of the network, thus allowing the fusion module to solely focus on feature selection. During inference, the modality-specific decoder heads are not used and the fused result is provided by the joint segmentation head. In the case of models that process each modality individually, the number of additional heads must be increased to match the number of all used modalities.

\subsection{Loss}
In contrast to the standard cross-entropy loss used regularly for semantic segmentation problems, which is calculated from model predictions and ground truth annotations, we can exploit both forward passes to more precisely guide the training process. In this section, the model predictions will be written as follows: $\mathcal{Z}_{\mathcal{ITL}}=\mathcal{Z}(\cdot,\cdot,\cdot)$ describes the model prediction using all three modalities ($\mathcal{I}$mage $\rightarrow$ RGB, $\mathcal{T}$hermal, $\mathcal{L}$IDAR), while $\mathcal{Z}_{\mathcal{I}}$ and $\mathcal{Z}_{\mathcal{A}}$ describe the predictions of modality-specific prediction heads for RGB and auxiliary data respectively. Cross-entropy loss will be written as $\mathcal{L}_{\mathcal{CE}}(\cdot, \cdot)$, while ground truth annotations will be written as $\mathcal{Z}_{\mathcal{GT}}$.

The final loss function is constructed as follows:
\begin{equation}
    \mathcal{L} = \mathcal{L}_{f} + \mathcal{L}_{s},
\end{equation}
where $\mathcal{L}_{f}$ and $\mathcal{L}_{s}$ denote the losses produced in the first and second forward pass, respectively. The first forward pass includes all the available data, its corresponding loss contribution being:
\begin{equation}
    \begin{aligned}
        \mathcal{L}_{f} = \mathcal{L}_{\mathcal{CE}}(\mathcal{Z}_{\mathcal{GT}}, \mathcal{Z}_{\mathcal{ITL}}) \\
        + \mathcal{L}_{\mathcal{CE}}(\mathcal{Z}_{\mathcal{GT}}, \mathcal{Z}_{\mathcal{I}}) \\
        + \mathcal{L}_{\mathcal{CE}}(\mathcal{Z}_{\mathcal{GT}}, \mathcal{Z}_{\mathcal{A}})
    \end{aligned}
\end{equation}
The second pass loss is similarly defined as:
\begin{equation}
    \begin{aligned}
        \mathcal{L}_{s} = \mathcal{L}_{\mathcal{CE}}(\mathcal{Z}_{\mathcal{GT}}, \mathcal{Z}(\varnothing, \mathcal{X}_{\mathcal{T}}, \mathcal{X}_{\mathcal{L}})) \\
        + \mathcal{L}_{\mathcal{CE}}(\mathcal{Z}_{\mathcal{GT}}, \mathcal{Z}_{\mathcal{A}}),
    \end{aligned}
\end{equation}
where $\mathcal{Z}(\varnothing, \mathcal{X}_{\mathcal{T}}, \mathcal{X}_{\mathcal{L}})$ denotes the model prediction when RGB input is set to zero.
The training process thus calculates the classical cross-entropy loss for each of the prediction heads in the first forward pass. The second forward pass then produces another prediction without using RGB information, which is again guided towards the ground truth annotations along with the auxiliary head prediction. Since the RGB information is not used in the second pass, the output of the RGB specific decoder head is ignored.

\section{Experiments}
\label{sec:experiment}

We perform the experiments that show the effect our changes have on existing multimodal architectures on our MULTIAQUA dataset, using the day-night data split. The modalities used are RGB images, thermal images and LIDAR point clouds. We deem these modalities to be the most potentially useful for different difficult circumstances on the water surface in poor lighting conditions. For daytime data, past approaches have shown (c.f. benchmark in~\cite{zust2023lars}) that RGB images can be used for precise predictions even in cluttered environments. Our aim was to explore the usability of other modalities that can take over as the main source of information when color images are no longer sufficient.

In our work, we evaluated three multimodal architectures that support an arbitrary number of additional modalities (CMNeXt~\cite{zhang2023delivering}, MMSFormer~\cite{reza2024mmsformer}, and StitchFusion~\cite{li2024stitchfusion}) on our dataset, specifically for the case of daytime and nighttime scene interpretation. We trained all architectures using \hspace*{0pt}Mix-Transformer backbones~\cite{xie2021segformer} pretrained on ImageNet. In the case of CMNeXt and MMSFormer, the non-RGB branches were initialized using the Kaiming initialization method~\cite{he2015delving}. StitchFusion uses shared backbone weights for all modalities, thus ImageNet pretrained weights were used. Following the CMNeXt implementation, the input data to the models were RGB images, thermal images, and LIDAR projections of shape $I \in \mathbb{R}^{h \times w \times 3}$ For single channel images, all three channels contain the same values, and for LIDAR, the channel values are the points’ distances.


For further insight into the performance of different methods and modalities, we also performed a modality-based ablation study. This was performed by evaluating the trained models on the nighttime test data, but not including all the modalities for inference. This can show how different modalities contribute to the final results and gives a clearer view of the usability of each of the modalities under difficult circumstances.


\subsection{MULTIAQUA day-night experiments}

The experimental evaluation of our proposed approach was performed on the \textit{day-night} split of our dataset MULTIAQUA. The main goal was to enable as much nighttime scene interpretation as possible, while retaining good performance on daytime images. We evaluated our approach on three multimodal architectures, all of them able to process an arbitrary number of additional modalities.

As stated in Section~\ref{sec:day-night}, the training and validation subsets were exclusively sampled from daytime data. The test set was constructed from nighttime data and served to challenge the model's ability to work in very poor lighting conditions. The validation set and the test set were used for gauging the model performance on daytime and nighttime data, respectively, since a robust model should be able to handle very difficult circumstances without that affecting its ability to perform in normal daytime situations. We report the mIoU metric for both validation (daytime) and test (nighttime) subsets. For test data, we separately report the IoU for the class \textit{dynamic obstacle}, due to it being the most difficult and the most crucial of the classes for ensuring the safety of nighttime navigation.

\begin{figure*}[ht]
  \centering
  \captionsetup{type=figure}
  \includegraphics[width=\linewidth]{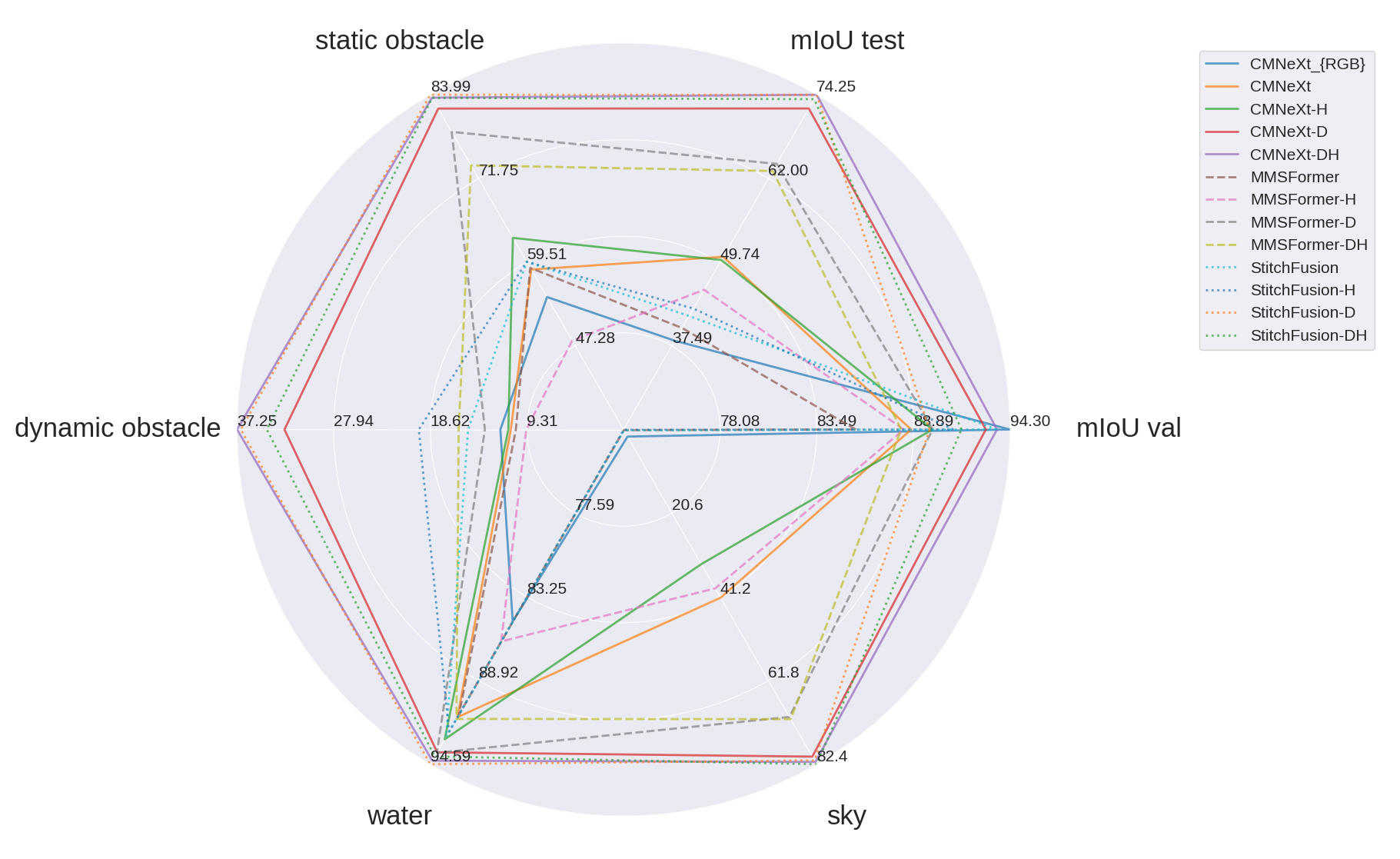}
  \caption{Radar chart of the experimental results on MULTIAQUA dataset. CMNeXt, MMSFormer, and StitchFormer variants are shown with solid, dashed lines, and dotted lines respectively. The scores shown are mIoU scores on validation and test sets, as well as IoU performances per semantic class (on the test set). The ordering of method variants matches the one in Table~\ref{tab:results}. The value range of each axis is scaled based on the corresponding data for clarity.}
    \label{fig:radar}
\end{figure*}

We trained multiple variants of each model, while introducing our proposed changes to the architecture, as can be observed in Table~\ref{tab:results}. For reference, we also show the RGB-only performance in the first row as CMNeXt$_{RGB}$. This variant is the CMNeXt architecture with no additional modalities, and is thus equivalent to the standard SegFormer architecture. We report the mean IoU scores across all classes for both the daytime validation and nighttime test subsets, as well as the IoU score specifically for the class \textit{dynamic obstacle} on the test set, which we deem the most important for safe navigation.

\begin{table*}[]
\resizebox{\textwidth}{!}{%
\begin{tabular}{c||cc||c|c|c}
\textbf{model} & \textbf{double pass} & \textbf{multihead} & \textbf{mIoU (val)} & \textbf{mIoU (test)} & \textbf{obstacle (test)} \\
\hhline{=#==#=|=|=}
CMNeXt$_{RGB}$ & ✗ & ✗ & \medal1{94.31} & 38.23 & 11.87 \\ \hline
CMNeXt & ✗ & ✗ & 88.78 & 50.52 & 10.86 \\ \hline
CMNeXt-H & ✗ & ✔ & 89.97 & 50.05 & 11.10 \\ \hline
CMNeXt-D & ✔ & ✗ & 92.95 & 72.24 & 32.68 \\ \hline
CMNeXt-DH & ✔ & ✔ & \medal2{93.58} & \medal1{74.25} & \medal1{37.25} \\ \hline
MMSFormer & ✗ & ✗ & 85.68 & 40.11 & 10.37 \\ \hline
MMSFormer-H & ✗ & ✔ & 88.36 & 45.70 & 9.37 \\ \hline
MMSFormer-D & ✔ & ✗ & 89.98 & 64.12 & 13.36 \\ \hline
MMSFormer-DH & ✔ & ✔ & 88.21 & 63.09 & 15.89 \\ \hline
StitchFusion & ✗ & ✗ & \medal3{93.57} & 41.86 & 14.98 \\ \hline
StitchFusion-H & ✗ & ✔ & 91.14 & 42.95 & 19.73 \\ \hline
StitchFusion-D & ✔ & ✗ & 89.81 & \medal2{74.23} & \medal2{36.89} \\ \hline
StitchFusion-DH & ✔ & ✔ & 91.59 & \medal3{73.59} & \medal3{34.38} \\ \hline
\end{tabular}%
}
\caption{Experimental results on MULTIAQUA validation (daytime) and test (nighttime) subsets. Modifications we introduced are also shown in the table. The last column contains the IoU metric for a single class \textit{dynamic obstacle} on the test set. The models are denoted with \textit{RGB} if only RGB data was used for training and inference, otherwise RGB, thermal and LIDAR data was used. The first, second and third-highest values per column are depicted in gold, silver and bronze, respectively.}
\label{tab:results}
\end{table*}

We denote different versions of models used in our experiment as follows: the suffix \textit{H} denotes multiple decoder heads, the suffix \textit{D} denotes a double pass training scheme, and the suffix \textit{DH} denotes the use of both. If no suffix is used, the baseline implementation and training protocol were used. The scores for different model variants are shown in Table~\ref{tab:results}. It can be observed that the baseline methods perform poorly on the nighttime test set, showing that the architectures were not able to extract meaningful information from the auxiliary modalities. Adding modality-specific decoder heads has different effects on the architecture performance. In the case of CMNeXt and MMSFormer, the results are marginally improved, while StitchFusion achieves a noticeable increase in obstacle detection, albeit at a cost of validation set performance.


As can be observed in Table~\ref{tab:results} and in Figure~\ref{fig:radar}, all the baseline models performed relatively well on daytime data, which is expected. However, the overall highest mIoU score on the validation set was achieved by the CMNeXt model only using RGB images. The performance on the test set was quite poor for the basic model version, none of them reaching a mIoU above 50. Introducing two forward passes induced the largest performance improvements. All the tested models' nighttime performance increased drastically, at more than 20 percentage points. The performance on the \textit{dynamic obstacle} class increased proportionally.

\begin{table*}[h]
\resizebox{\textwidth}{!}{%
\begin{tabular}{l||cccccc}

& \textbf{DeLiVER}~\cite{zhang2023delivering}  & DeLiVER (fog) & DeLiVER (night) & \textbf{MCubeS}~\cite{Liang_2022_CVPR}  & \textbf{MFNet}~\cite{ha2017mfnet}  & \textbf{Freiburg Thermal}~\cite{vertens2020heatnet}  \\
\hhline{=#======}
CMNeXt~\cite{zhang2023delivering}   & \medal2{47.11} & 47.22 & 44.64 & \medal2{44.92} & \medal3{49.53} & \medal3{20.69} \\ \hline
CMNeXt-H  & \medal1{48.69} & \medal1{57.77} & \medal1{56.01} & \medal1{46.08} & \medal2{51.37} & 17.4 \\ \hline
CMNeXt-D  & 42.76 & \medal3{52.13} & \medal3{52.38} & 42.09 & 46.23 & \medal2{34.02} \\ \hline
CMNeXt-DH & \medal3{45.99} & \medal2{56.71} & \medal2{55.46} & \medal3{42.52} & \medal1{51.78} & \medal1{35.39} \\ \hline
\end{tabular}
}
\caption{Segmentation results on different multimodal datasets. All available modalities were used for all datasets, namely: DeLiVER (RGB, depth, event, LIDAR), MCubeS (RGB, AoLP, DoLP, NIR), MFNet (RGB, thermal), HeatNet (RGB, thermal). For the DeLiVER dataset, we also report the performance on adverse condition subsets \textit{fog} and \textit{night}. The first, second and third-highest values per column are depicted in gold, silver and bronze, respectively.}
\label{tab:extra_datasets}
\end{table*}

Introducing modality-specific decoder heads had a mixed response, where CMNeXt's performance increased in both single and double pass regimes, whereas this could only be observed in single pass regime for MMSFormer and StitchFusion. However, just introducing multiple decoder heads to StitchFusion improved the \textit{dynamic obstacle} IoU for almost 5 percentage points on its own. Another interesting observation is that our modifications cause the CMNeXt model to also perform better on daytime data, while increasing the nighttime performance. This is also true for MMSFormer, albeit by a smaller extent. StitchFusion achieves the second-highest nighttime performance using our modifications, but at the cost of its daytime performance. Overall, the CMNeXt achieved the highest overall scores of all multimodal architectures using our modifications, improving both daytime and nighttime performance. Given that CMNeXt only using color images performed best on daytime data, there is still room for improvement.

\subsection{Other datasets experiments}

We also trained the best-performing method on other multimodal datasets to ascertain if our approach could improve performance in general, or is the performance gain limited to our dataset. We trained the different CMNeXt variants on datasets DELIVER~\cite{zhang2023delivering}, MCubeS~\cite{Liang_2022_CVPR}, MFNet~\cite{ha2017mfnet} and Freiburg Thermal~\cite{vertens2020heatnet}. The datasets used are well established multimodal datasets, including different modalities and focusing on semantic segmentation for autonomous driving, pedestrian detection and material segmentation. DELIVER dataset is synthetic, whereas the others are not, and none of them contains maritime scenarios, thus they are very different in appearance and structure from our proposed dataset. The Freiburg Thermal dataset is conceptually closest to our dataset, since it also includes daytime and nighttime data, where nighttime data is only annotated for the testing subset, but it was captured using an autonomous ground vehicle. We only trained the models on daytime data from Freiburg Thermal dataset and tested only on nighttime data, same as with MULTIAQUA. The results are shown in Table~\ref{tab:extra_datasets}. The backbone used was MiT-B0, as opposed to the main experiment, which uses MiT-B4. This was purely because of hardware limitations. Our results do not represent state-of-the-art performance, but only relative changes induced by our model and training modifications, to test their generality. It can be observed that training the model using our method increases the performance on all tested datasets. The manner in which our approach benefits datasets seems to be tied to the structure and purpose of the datasets. Both DeLiVER and MCubeS use test sets that are relatively similar to the training data, whereas MFNet contains nighttime data in both training and testing sets, and Freiburg Thermal was trained and tested similarly to MULTIAQUA. Freiburg Thermal thus also shows similar score changes as Table~\ref{tab:results}. On DeLiVER and MCubeS, the highest score is obtained using CMNeXt-H variant that only uses modality-specific decoder heads, and the other datasets benefit the most by using both our modifications. It can also be observed that the effect of our modifications is even more pronounced when evaluated on adverse weather conditions in the DeLiVER dataset (the second and third column show the results for \textit{fog} and \textit{night} subsets). The increase in mIoU results in the case of CMNeXt-H under difficult conditions was over 10 percentage points relative to the baseline model. The changes induced are not nearly as large as on MULTIAQUA, but show the potential for our modifications to benefit the model performance in general, not just on our dataset and domain.


\subsection{Modality-based ablation study}

We performed a modality-based ablation study to further evaluate the performance of the trained models. Usually ablation studies are performed by changing or removing parts of the architecture, but in our case, evaluating the models only using partial data seemed reasonable. This process shows how well the models are able to handle missing data and produce quality predictions despite that. Every model was evaluated on the MULTIAQUA nighttime test set, each time with a different subset of data modalities. This consisted of using each modality on its own and of all possible pairs of two modalities. The results are shown in Table~\ref{tab:ablation}. The scores reported for each data split of each model are the mIoU over all classes, then the IoU of the \textit{dynamic obstacle} class. It can be observed that the performance of different data splits varies wildly between the models. The main observation is that models that generally perform well on the nighttime data appear to be able to produce quality predictions using only the available modalities. The top-performing models' test scores should monotonically increase when adding more modalities, showing that the model can utilize new information. The score should never decrease when more data is available at inference time. It can be observed this is not entirely true at the moment, with several model versions performing better on nighttime data if RGB data is not present at all.

\begin{table*}[]
\resizebox{\textwidth}{!}{%
\begin{tabular}{|c|cc|cc|cc|cc|cc|cc|cc|}
\hline
 & \multicolumn{2}{c|}{img} & \multicolumn{2}{c|}{thermal} & \multicolumn{2}{c|}{LIDAR} & \multicolumn{2}{c|}{img, thermal} & \multicolumn{2}{c|}{img, LIDAR} & \multicolumn{2}{c|}{thermal, LIDAR} & \multicolumn{2}{c|}{img, thermal, LIDAR} \\ \hline
 model & \multicolumn{1}{c|}{mIoU} & obstacle & \multicolumn{1}{c|}{mIoU} & obstacle & \multicolumn{1}{c|}{mIoU} & obstacle & \multicolumn{1}{c|}{mIoU} & obstacle & \multicolumn{1}{c|}{mIoU} & obstacle & \multicolumn{1}{c|}{mIoU} & obstacle & \multicolumn{1}{c|}{mIoU} & obstacle \\ \hline
CMNeXt & \multicolumn{1}{c|}{\medal1{50.7}} & 7.75 & \multicolumn{1}{c|}{30.8} & 0.0 & \multicolumn{1}{c|}{41.4} & 0.0 & \multicolumn{1}{c|}{\medal3{50.41}} & 7.74 & \multicolumn{1}{c|}{49.37} & 9.04 & \multicolumn{1}{c|}{37.49} & 0.0 & \multicolumn{1}{c|}{\medal2{50.52}} & 10.86 \\ \hline
CMNeXt-H & \multicolumn{1}{c|}{46.19} & 7.77 & \multicolumn{1}{c|}{37.34} & 0.0 & \multicolumn{1}{c|}{38.16} & 0.0 & \multicolumn{1}{c|}{\medal2{49.22}} & 9.09 & \multicolumn{1}{c|}{\medal3{48.05}} & 9.86 & \multicolumn{1}{c|}{34.81} & 0.0 & \multicolumn{1}{c|}{\medal1{50.05}} & 11.1 \\ \hline
CMNeXt-D & \multicolumn{1}{c|}{52.03} & 0.76 & \multicolumn{1}{c|}{59.56} & 7.23 & \multicolumn{1}{c|}{\medal3{61.15}} & 18.28 & \multicolumn{1}{c|}{60.18} & 8.09 & \multicolumn{1}{c|}{59.71} & 17.33 & \multicolumn{1}{c|}{\medal1{73.14}} & 35.11 & \multicolumn{1}{c|}{\medal2{72.24}} & 32.68 \\ \hline
CMNeXt-DH & \multicolumn{1}{c|}{36.17} & 4.56 & \multicolumn{1}{c|}{63.66} & 17.54 & \multicolumn{1}{c|}{52.28} & 21.9 & \multicolumn{1}{c|}{\medal3{63.83}} & 17.41 & \multicolumn{1}{c|}{50.81} & 21.35 & \multicolumn{1}{c|}{\medal1{74.45}} & 38.06 & \multicolumn{1}{c|}{\medal2{74.25}} & 37.25 \\ \hline
MMSFormer & \multicolumn{1}{c|}{38.58} & 5.38 & \multicolumn{1}{c|}{26.91} & 0.0 & \multicolumn{1}{c|}{26.82} & 0.0 & \multicolumn{1}{c|}{\medal3{38.61}} & 5.48 & \multicolumn{1}{c|}{\medal2{40.1}} & 10.37 & \multicolumn{1}{c|}{26.8} & 0.0 & \multicolumn{1}{c|}{\medal1{40.11}} & 10.37 \\ \hline
MMSFormer-H & \multicolumn{1}{c|}{34.09} & 7.34 & \multicolumn{1}{c|}{33.36} & 0.0 & \multicolumn{1}{c|}{\medal3{36.4}} & 0.0 & \multicolumn{1}{c|}{\medal2{45.0}} & 8.65 & \multicolumn{1}{c|}{34.8} & 8.1 & \multicolumn{1}{c|}{33.7} & 0.0 & \multicolumn{1}{c|}{\medal1{45.7}} & 9.37 \\ \hline
MMSFormer-D & \multicolumn{1}{c|}{49.99} & 3.72 & \multicolumn{1}{c|}{\medal3{61.35}} & 0.12 & \multicolumn{1}{c|}{39.43} & 0.24 & \multicolumn{1}{c|}{60.52} & 4.26 & \multicolumn{1}{c|}{40.12} & 5.11 & \multicolumn{1}{c|}{\medal1{67.91}} & 14.05 & \multicolumn{1}{c|}{\medal2{64.12}} & 13.36 \\ \hline
MMSFormer-DH & \multicolumn{1}{c|}{41.54} & 5.65 & \multicolumn{1}{c|}{51.31} & 0.8 & \multicolumn{1}{c|}{42.86} & 0.03 & \multicolumn{1}{c|}{\medal3{59.36}} & 7.38 & \multicolumn{1}{c|}{42.76} & 9.3 & \multicolumn{1}{c|}{\medal2{59.92}} & 4.89 & \multicolumn{1}{c|}{\medal1{63.09}} & 15.89 \\ \hline
StitchFusion & \multicolumn{1}{c|}{38.55} & 5.05 & \multicolumn{1}{c|}{38.65} & 0.0 & \multicolumn{1}{c|}{37.74} & 0.72 & \multicolumn{1}{c|}{39.6} & 6.72 & \multicolumn{1}{c|}{\medal2{40.3}} & 11.76 & \multicolumn{1}{c|}{\medal3{39.42}} & 2.73 & \multicolumn{1}{c|}{\medal1{41.86}} & 14.98 \\ \hline
StitchFusion-H & \multicolumn{1}{c|}{39.19} & 6.34 & \multicolumn{1}{c|}{31.1} & 0.09 & \multicolumn{1}{c|}{25.49} & 0.0 & \multicolumn{1}{c|}{\medal2{40.89}} & 11.55 & \multicolumn{1}{c|}{\medal3{40.43}} & 11.46 & \multicolumn{1}{c|}{32.62} & 15.22 & \multicolumn{1}{c|}{\medal1{42.95}} & 19.73 \\ \hline
StitchFusion-D & \multicolumn{1}{c|}{39.32} & 0.12 & \multicolumn{1}{c|}{65.87} & 12.91 & \multicolumn{1}{c|}{61.18} & 17.86 & \multicolumn{1}{c|}{\medal3{66.16}} & 13.45 & \multicolumn{1}{c|}{61.89} & 21.4 & \multicolumn{1}{c|}{\medal1{74.89}} & 38.87 & \multicolumn{1}{c|}{\medal2{74.23}} & 36.89 \\ \hline
StitchFusion-DH & \multicolumn{1}{c|}{40.59} & 3.06 & \multicolumn{1}{c|}{64.42} & 6.61 & \multicolumn{1}{c|}{52.65} & 25.01 & \multicolumn{1}{c|}{\medal3{66.44}} & 14.07 & \multicolumn{1}{c|}{50.36} & 20.36 & \multicolumn{1}{c|}{\medal1{74.51}} & 37.01 & \multicolumn{1}{c|}{\medal2{73.59}} & 34.38 \\ \hline
\end{tabular}%
}
\caption{Segmentation results for the modality-based ablation study. Each of the trained models was evaluated with every combination of input modalities. Shown are the mIoU scores on the nighttime test set of MULTIAQUA and the IoU scores for the class \textit{dynamic obstacle}. The first, second and third-highest values in every row are depicted in gold, silver and bronze, respectively.}
\label{tab:ablation}
\end{table*}

\subsection{Implementation details}
We trained models using MULTIAQUA data at resolution 1152$\times$640 px, which is half the resolution of ZED images, and used a batch size of 4. The DeLiVER dataset was trained at resolution 1024$\times$1024 px with batch size 1, MCubeS was trained at resolution 512$\times$512 px with batch size 2, MFNet was trained at resolution 640$\times$480 px with batch size 4, while Freiburg Thermal was trained at resolution 960$\times$325 px with batch size 4. The different setups were used due to VRAM constraints, but it should be noted that the experiments performed were used to gauge the performance of different model variants, not to obtain competitive performance. That would require complex backbone settings, fine-tuning the hyperparameters and a larger number of training epochs.
All the models trained in the experimental section were trained using the MiT-B4 backbone~\cite{xie2021segformer}, except for the models trained on datasets other than MULTIAQUA, which were trained on MiT-B0, to constrain the memory footprint and training time. The relative comparisons should however still hold even for more complex MiT backbones. We trained all models for 100 epochs using a learning rate of 1e-5. For evaluation, we used the model that performed best on the corresponding validation or test subset.

In Table~\ref{tab:memory} we report the number of parameters and GFLOPS of different variants of models used in our experiments. We can observe that the increased number of parameters is caused mostly by adding modality-specific decoder heads. Furthermore, the number of operations increases significantly, since despite their simple design, the decoder heads carry out a large number of operations. This is however bounded by the architecture design in the case of CMNeXt, since only one decoder head is added for each of the RGB and auxiliary branches, thus the memory footprint only increases by a constant amount. This however does not hold for MMSFormer architecture, where every extra modality incurs additional memory requirements. StitchFusion has a lower memory footprint by default, due to its weight sharing architecture. However, adding more auxiliary modalities would increase the complexity of the model, because modality-pair MultiAdapter modules must be added for all pairs of modalities.

\begin{table}[!htb]
\resizebox{\columnwidth}{!}{
\begin{tabular}{c||c|c}
model & \# of parameters [M] & GFLOPS \\
\hhline{=#=|=}
CMNeXt$_{RGB}$ & 62 & 72 \\ \hline
CMNeXt & 117 & 123 \\ \hline
CMNeXt-H & 120 & 159 \\ \hline
CMNeXt-D & 117 & 123 \\ \hline
CMNeXt-DH & 120 & 159 \\ \hline
MMSFormer & 187 & 181 \\ \hline
MMSFormer-H & 191 & 236 \\ \hline
MMSFormer-D & 187 & 181 \\ \hline
MMSFormer-DH & 191 & 236 \\ \hline
StitchFusion & 66 & 186 \\ \hline
StitchFusion-H & 71 & 241 \\ \hline
StitchFusion-D & 66 & 186 \\ \hline
StitchFusion-DH & 71 & 241
\end{tabular}%
}
\caption{Number of parameters and GFLOPS for different variants of CMNeXt, MMSFormer and StitchFusion models used in our experiments.}
\label{tab:memory}
\end{table}

\section{Conclusions}

In this paper, we presented a publicly available multimodal maritime dataset MULTIAQUA that includes several cameras along with LIDAR, radar and GPS data. We also provide per-pixel semantic annotations that can be used for supervised learning. A challenging nighttime testing subset is included to evaluate the performance in low-light scenarios. We also presented modifications to existing multimodal architectures and training protocol which can be used to improve model performance in difficult circumstances. Specifically, we focused on the case of training a multimodal method only using daytime data, and showed that our approach greatly improves performance on our difficult nighttime test set. Our results show that a model trained using our approach can use thermal images and LIDAR data to consistently perform semantic segmentation even if RGB data is poor or unavailable. Since the training data did not include any nighttime samples, we expect that this approach can be generalized to other difficult situations, such as fog, rain and snow, as long as least \emph{some} of the sensors in our multimodal setup remain largely unaffected by the conditions.

We trained variants of three multimodal models on our dataset and showed that our approach achieves greatly improved performance on nighttime data relative to baseline methods. We also showed that our approach could improve performance on other multimodal datasets as well, and performed a modality-ablation study to show that the models trained using our approach are able to handle missing data better than baseline models.

Our proposed dataset can be used to train sensor-specific models or include additional modalities in a multimodal architecture, which could improve the performance in both daytime and nighttime scenarios even further. Polarization camera or NIR camera data could be used to improve separation between water and shore and radar could be used to verify obstacle detections in poor visibility, harsh sunlight or fog.

Additionally, our proposed method could be improved by more precisely modeling possible data degradations by employing different data degradation schemes and possibly a module for data quality detection on each modality branch to predict how useful a modality is for final prediction. Such a module could be applied to all modalities, and improve the model's understanding of each sensor's data quality, so fusion and final predictions could be performed only using the most informative modalities.


%





\ifCLASSOPTIONcaptionsoff
  \newpage
\fi



\bibliographystyle{IEEEtran}
\bibliography{bibliography}
\end{document}